\documentclass[10pt,twocolumn,letterpaper]{article}

\usepackage{cvpr}
\usepackage{times}
\usepackage{epsfig}
\usepackage{graphicx}
\usepackage{amsmath}
\usepackage{amssymb}

\usepackage{blindtext}
\usepackage{bbm}
\usepackage{enumitem}
\usepackage{epsfig}
\usepackage[ruled]{algorithm2e}
\usepackage{algorithmic}
\usepackage{multirow}
\usepackage{color}
\usepackage{xcolor}
\usepackage{color, colortbl}
\usepackage{array}
\usepackage[english]{babel} 
\usepackage{amsmath,amsfonts,amsthm} 
\usepackage{mathtools}
\usepackage{pifont}
\usepackage{authblk}

\usepackage{threeparttable}
\usepackage{subfigure}
\usepackage{booktabs}
\usepackage{appendix}
\expandafter\def\expandafter\normalsize\expandafter{%
	\normalsize\setlength\abovedisplayskip{3pt}}
\expandafter\def\expandafter\normalsize\expandafter{%
	\normalsize\setlength\belowdisplayskip{3pt}}

\usepackage[pagebackref=true,breaklinks=true,letterpaper=true,colorlinks,bookmarks=false]{hyperref}

\cvprfinalcopy 

\def\cvprPaperID{188} 

\ifcvprfinal\fi
\begin{document}
	
	\title{Supervised Auto-Encoding Twin-Bottleneck Hashing}
	\author[1]{Yuan Chen\thanks{Corresponding Author}
	}
	\author[1]{St\'ephane Marchand-Maillet
	}
	
	\vspace{-4ex}
	\affil[1]{University of Geneva, Geneva, Switzerland}
	\affil[ ]{\small \texttt{ychen169@ucr.edu}}
	
	\maketitle
	\thispagestyle{empty}
	
	\begin{abstract}
		Deep hashing has shown to be a complexity-efficient solution for the Approximate Nearest Neighbor search problem in high dimensional space. Many methods usually build the loss function from pairwise or triplet data points to capture the local similarity structure. Other existing methods construct the similarity graph and consider all points simultaneously. Auto-encoding Twin-bottleneck Hashing is one such method that dynamically builds the graph. Specifically, each input data is encoded into a binary code and a continuous variable, or the so-called twin bottlenecks. The similarity graph is then computed from these binary codes, which get updated consistently during the training. In this work, we generalize the original model into a supervised deep hashing network by incorporating the label information. In addition, we examine the differences of codes structure between these two networks and consider the class imbalance problem especially in multi-labeled datasets. Experiments on three datasets yield statistically significant improvement against the original model. Results are also comparable and competitive to other supervised methods.
	\end{abstract}

	\section{Introduction}\label{sec:intro}
	
	In the age of information explosion, storing massive data and retrieving relevant samples from the dataset in high dimensional space are fundamental topics in both the academic and industrial communities. Learning to hash is one promising solution and has gained lots of attention. The hashing method aims to map complex input data into compact binary codes, which respect the local similarity structure of the original space. This way, time-consuming distance computation in the Euclidean space can be approximately replaced by the simple hamming distance computation in the binary space, which can be performed efficiently in the hardware using the CPU instruction \cite{wang2017survey}. In addition, storing binary codes instead of original data also reduces the burden on the storage requirement.
	
	Existing hashing methods can be roughly divided into traditional methods and deep hashing. Traditional methods \cite{weiss2008spectral}\cite{liu2012supervised}\cite{wang2013learning}\cite{gong2012iterative}\cite{jegou2010product}\cite{zhang2014composite} take feature vectors as input and engineer a hash function with parameters, which can be optimized by a loss function. In contrast, deep hashing \cite{liu2016deep}\cite{do2016learning}\cite{li2015feature}\cite{li2017deep}\cite{shen2020auto} takes advantage of neural networks to extract features and plays the role of hash function thanks to its powerful ability of simulating complicated functions. The network parameters are optimized by the loss function, and this strategy has demonstrated superior performance over the traditional methods. Deep methods can also again fall into unsupervised methods \cite{liu2016deep}\cite{shen2020auto}\cite{zhu2016deep} and supervised methods \cite{do2016learning}\cite{li2017deep}\cite{wang2016deep}\cite{li2015feature}\cite{peng2019deep}\cite{yuan2018relaxation}\cite{cakir2018hashing}\cite{wang2020deep}.
	
	Since the hash function needs to preserve the neighborhood structure of the input space, existing methods usually involve similarity terms computed from the pair or triplet of data points, or construct the similarity graph. Jie Qin et al. \cite{shen2020auto} pointed out that these methods could suffer from the "static graph" problem where the pre-computed graph would bring biased prior knowledge and could not be updated to capture the intrinsic data structure, leading to sub-optimal performance. To address the problem, Twin-Bottleneck Hashing (TBH) \cite{shen2020auto} introduces twin bottlenecks: the binary bottleneck (BinBN) and the continuous bottleneck (ConBN). The similarity graph is constructed directly from the binary codes and gets updated during the training process to better model the data structure. In addition, the ConBN could help improve the reconstruction quality in the decoder, and the reconstruction loss of the decoder is optimized as a score to quantify the encoder quality. However, minimizing the reconstruction loss only provides an implicit guide/direction on optimizing the binary codes, and the back-propagation pathway flows from the decoder to the GCN network \cite{kipf2016semi} before reaching the encoder. In other words, the binary codes may not receive adequate guidance to be updated. 
	On the other hand, class labels are not considered in the model, and the architecture is too weak to operate self-supervision. In fact, label information is important for the hashing problem, since it acts as an independent description for data points besides the information captured by the feature vectors, which complements to provide a more complete view of the same object and help encode binary codes of better quality.
	
	As a solution to the above problems, this paper proposes the Supervised Twin-Bottleneck Hashing (STBH). Based on the original TBH model, we introduce a classification layer built on top of the binary bottleneck to incorporate the semantic information. This brings a number of benefits. First, the binary codes receive direct optimizing direction from the classification loss term. Second, supervised information mitigates the semantic gap and makes the distinction of codes better separated between different classes. Third, the loss function also introduces robustness when the extracted features incorrectly reflect the distance relation with respect to its neighbors. The main contributions of this work are summarized as follows:
	\begin{itemize}
		\item A classification layer is proposed on top of the binary bottleneck of the TBH model to include the label information.
		\item We study and compare the codes structures between the two networks to show the benefits brought by the supervised learning.
		\item We consider the class imbalance problem in multi-label datasets, which are hardly mentioned previously.
		\item Experiments conducted on three datasets show superior performances compared to the original model.
	\end{itemize}
	
	\section{Related Work}\label{sec:related}
	
	The proposed STBH is mostly related to the TBH model \cite{shen2020auto}, and thus inherits the same competing techniques.
	
	TBH model \cite{shen2020auto} is one of unsupervised deep hashing with encoder-decoder architecture. A twin-bottleneck is incorporated to construct an adaptive similarity graph and mitigate the reconstruction loss of the decoder component. In addition, a stochastic neuron \cite{song2017stochastic} is utilized to allow the standard back-propagation process while preserving the binary constraints. In order to achieve the balanced and independent binary codes, the model involves Wasserstein Auto-Encoders (WAE) \cite{tolstikhin2017wasserstein} to regulate the binary variables adversarially. As discussed above, the binary codes are implicitly optimized by the reconstruction loss, and the absence of the semantic information could also lead to sub-optimal performance.
	
	\section{Proposed Method}
	
	\begin{figure*}[!ht]
		\center
		\includegraphics[width=\textwidth]{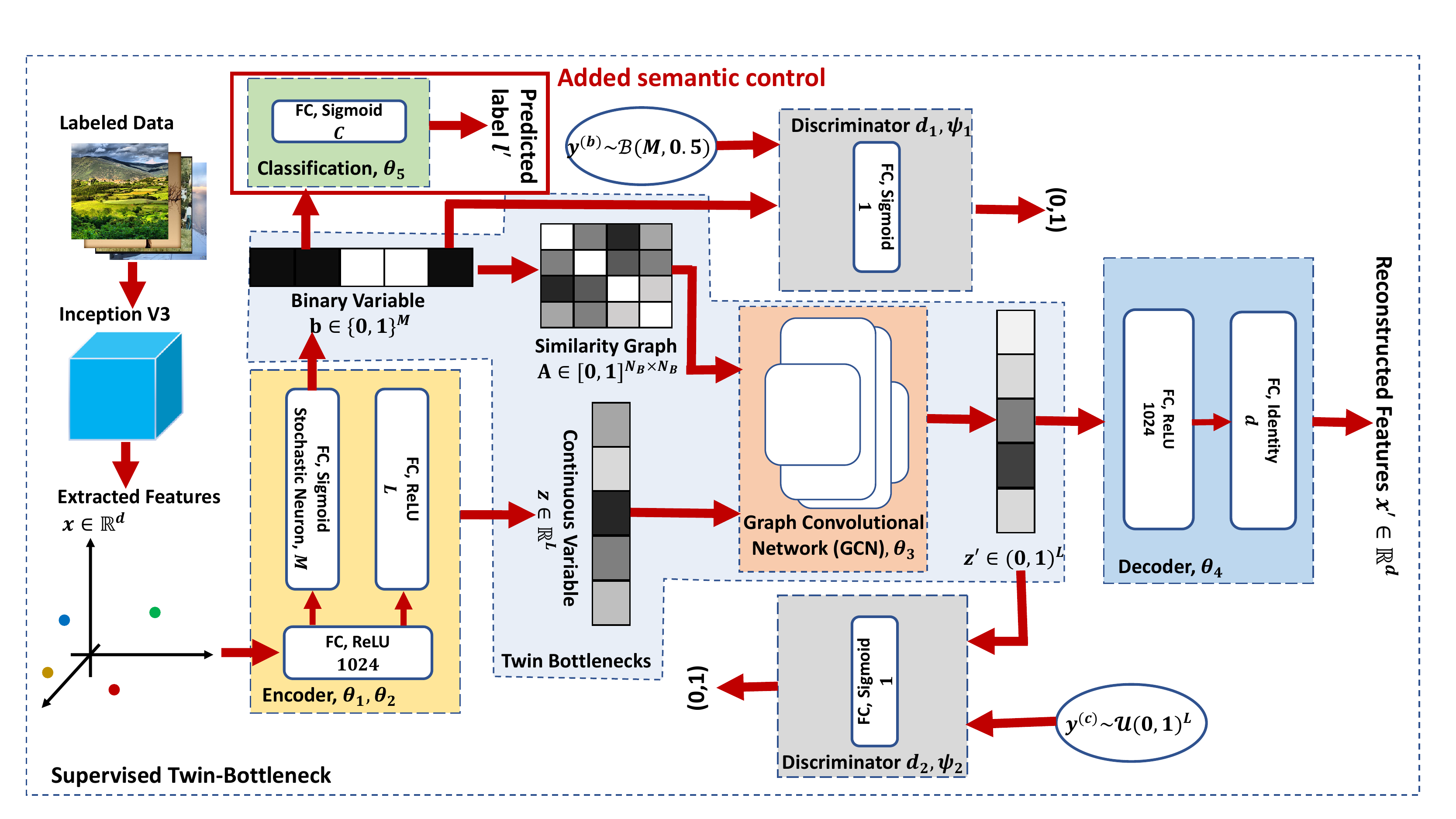}
		\caption{The schematic of STBH. A classification layer is directly added on top of the binary bottleneck to instill the semantic information. The number of classes is denoted as $C$. Other modules follow the pattern of the original TBH model.}
		\label{fig:stbh}
	\end{figure*}
	
	\subsection{Notations and TBH}
	
	Our STBH is established based on the unsupervised TBH model \cite{shen2020auto}. We first introduce notations as well as the TBH network briefly.
	
	The network structure is illustrated in Fig. \ref{fig:stbh}. First, feature vectors, denoted by $\mathbf{x} \in \mathbb{R}^{D}$, are extracted from the input raw data. Then the feature vector is fed to the encoder to generate two latent variables: a $M$-bit binary code $\mathbf{b}$, and a $L$ dimensional continuous variable $\mathbf{z}$. Specifically:
	\begin{equation}
		\begin{aligned}
			\mathbf{b} &= \alpha\left( f_{1}(\mathbf{x};\theta_{1}), \epsilon \right) \in \{0,1\}^{M},\\
			\mathbf{z} &= f_{2}\left( \mathbf{x};\theta_{2} \right) \in \mathbb{R}^{L}
		\end{aligned}
	\end{equation}
	where $f_{1}$ and $f_{2}$ indicate the two encoding functions, and $\theta_{1}$ and $\theta_{2}$ are the corresponding layer parameters of $f_{1}$ and $f_{2}$. In addition, to implement the standard back-propogation (BP) process with binary constraints,  the discrete stochastic neuron activation  $\alpha\left( \cdot,\epsilon \right)$ \cite{song2017stochastic} is utilized to produce binary codes and defined as:
	\begin{equation}
		\label{eqn:neuron}
		b^{i} = \alpha\left( f_{1}(\mathbf{x};\theta_{1}),\epsilon \right)^{i} = \begin{cases}
			&1 \quad f_{1}(\mathbf{x};\theta_{1})^{i} \geq \epsilon^{i},\\
			&0 \quad f_{1}(\mathbf{x};\theta_{1})^{i} < \epsilon^{i}
		\end{cases}
	\end{equation}
	where the superscript $i$ indicate the $i$-th component of the vector, and variables $\epsilon$ conform to the uniform distribution $\mathcal{U}(0,1)^{M}$.
	
	For simplicity, we consider a batch of $N_{B}$ data points, and denote corresponding continuous variables $\mathbf{Z}_{B}$ and binary codes $\mathbf{B}_{B}$ as: 
	\begin{equation}
		\begin{aligned}
			\mathbf{Z}_{B} &= [ \mathbf{z}_{1};\mathbf{z}_{2};\cdot;\mathbf{z}_{N_{B}} ] \in \mathbb{R}^{N_{B}\times L}\\
			\mathbf{B}_{B} &= [ \mathbf{b}_{1};\mathbf{b}_{2};\cdot;\mathbf{b}_{N_{B}} ] \in \{0,1\}^{N_{B}\times M}
		\end{aligned}
	\end{equation}
	Subsequently, the similarity graph $\mathbf{A}$ can be computed from binary codes of the training batch by:
	\begin{equation}
		\begin{aligned}
			\mathbf{A} & = \mathbb{J} + \frac{1}{M}\left( \mathbf{B}_{B}(\mathbf{B}_{B}-\mathbb{J})^{T}+(\mathbf{B}_{B}-\mathbb{J})\mathbf{B}_{B}^{T} \right) \\
			& \in [0,1]^{N_{B}\times N_{B}}
		\end{aligned}
	\end{equation}
	where $\mathbb{J}$ is a $N_{B}\times N_{B}$ matrix with all elements equal to 1. Later, the graph and the continuous variables are processed by the GCN network \cite{kipf2016semi} to generate the latent variables $\mathbf{Z}_{B}^{'}$ for reconstruction:
	\begin{equation}
		\begin{aligned}
			\mathbf{Z}_{B}^{'} & = \mathtt{sigmoid}\left( \mathbf{D}^{-\frac{1}{2}}\mathbf{A}\mathbf{D}^{-\frac{1}{2}}\mathbf{Z}_{B}\mathbf{W}_{\theta_{3}} \right)\\
			&  \in (0,1)^{N_{B}\times L}
		\end{aligned}
	\end{equation}
	where $\mathbf{W}_{\theta_{3}}$ is a $L \times L$ matrix of parameters of the GCN network. In addition,
	\begin{equation}
		\mathbf{D}=\mathtt{diag}\left( \mathbf{A}\mathbf{1}^{T} \right)
	\end{equation}
	where $\mathbf{1}$ is a $N_{B}$-dimensional vector full of 1s. Finally $\mathbf{Z}_{B}^{'}$ is fed into the decoder for the reconstruction of feature vectors:
	\begin{equation}
		\hat{\mathbf{x}} = g\left( \mathbf{z}';\theta_{4} \right) \in \mathbb{R}^{D}
	\end{equation}
	where $g(\cdot;\theta_{4})$ indicates the decoding function and $\theta_{4}$ is the corresponding parameters.
	
	In addition, to encourage latent variables to fully explore the latent spaces, two WAEs \cite{tolstikhin2017wasserstein} are utilized to adversarially regulate variables with two discriminators: $d_{1}\left( \cdot;\psi_{1} \right)$ and $d_{2}\left( \cdot;\psi_{2} \right)$ where $\psi_{1}$ and $\psi_{2}$ are corresponding parameters. Specifically, $d_{1}\left( \cdot;\psi_{1} \right)$ is used for the binary codes $\mathbf{b}$ to achieve the bit balance and independence, while $d_{2}\left( \cdot;\psi_{2} \right)$ is employed for the latent variables $\mathbf{z}^{'}$:
	\begin{equation}
		\begin{aligned}
			d_{1}\left( \mathbf{b};\psi_{1} \right) \in (0,1);&\quad d_{1}\left( \mathbf{y}^{b};\psi_{1} \right) \in (0,1)\\
			d_{2}\left( \mathbf{z}';\psi_{2} \right) \in (0,1);&\quad
			d_{2}\left( \mathbf{y}^{c};\psi_{2} \right) \in (0,1)
		\end{aligned}
	\end{equation}
	where $\mathbf{y}^{b}$ conforms to binomial distribution $\mathcal{B}(M, 0.5)$ and $\mathbf{y}^{c}$ is sampled from the uniform distribution $\mathcal{U}(0,1)^{L}$.
	
	\subsection{Supervised TBH}
	As shown in Fig. \ref{fig:stbh}, we propose to add a classification layer directly on top of the binary variables, which consists of one fully connected layer with sigmoid as its activation function. The predicted label $\mathbf{l}'$ can be expressed as:
	\begin{equation}
		\mathbf{l}' = \mathtt{sigmoid}\left( \mathbf{W}_{\theta_{5}}\mathbf{b} \right) \in [0,1]^{C}
	\end{equation}
	where $\mathbf{W}_{\theta_{5}}  \in \mathbb{R}^{C\times M}$ indicates the projection matrix and $C$ is the number of classes. This classifier can not only introduce the direct guidance on the binary codes, but also provide semantic information to make codes distribution of different classes more distinguishable, which could help improve the codes quality. We detail how this semantic control is incorporated into the loss function in the next section.
	
	\subsection{Loss Function}
	
	\subsubsection{Discrminating Objective}
	
	The Discrminating objective $\mathcal{L}_{D}$ is defined by:
	\begin{equation}
		\label{eqn:dis}
		\begin{aligned}
			\mathcal{L}_{D} &=  \frac{1}{N_{B}}\sum_{i=1}^{N_{B}}\left( \log{d_{1}( \mathbf{y}_{i}^{b};\psi_{1})} \right. \\
			& + \left. \log{\left(1-d_{1}(\mathbf{b}_{i};\psi_{1})\right)} \right. \\
			& + \left. \log{d_{2}( \mathbf{y}_{i}^{c};\psi_{2})} + \log{\left(1-d_{2}(\mathbf{z}_{i}^{'};\psi_{2})\right)} \right)
		\end{aligned}
	\end{equation}
	$\mathcal{L}_{D}$ is optimized to train two discriminators over the parameter space $\{ \psi_{1},\psi_{2} \}$ such that they could distinguish the distributions of codes generated by the encoder from those sampled from the targeted distributions.
	
	\subsubsection{Auto-Encoding Objective}
	
	The Auto-Encoding objective $\mathcal{L}_{AE}$ is defined as:
	\begin{equation}
		\label{eqn:enc}
		\begin{aligned}
			\mathcal{L}_{AE} & =  \frac{1}{N_{B}}\sum_{i=1}^{N_{B}}\mathbb{E}_{\mathbf{b}_{i}}\left[ \frac{1}{D}\|\mathbf{x}_{i}-\mathbf{x}_{i}^{'}\|^{2} \right.\\
			& + \lambda\log{d_{1}(\mathbf{b}_{i})} + \lambda\log{d_{2}(\mathbf{z}_{i}^{'})}\\
			& + \left. \gamma\|\mathbf{l}_{i}-\mathbf{l}_{i}^{'}\|^{2} \right]\\
			& + \eta\|\mathbf{W}_{\theta_{5}}\|_{1,1}
		\end{aligned}
	\end{equation}
	where $\mathbf{l}$ denotes the true label, $\lambda$, $\gamma$ and $\eta$ are hyper-parameters to control the weights of the discriminating loss, the regression loss and the regularization of classifier's weights respectively. $\mathbb{E}_{\mathbf{b}_{i}}$ means the expectation over the latent binary code $\mathbf{b}$ since $\mathbf{b}$ is generated from a sampling process. Specifically, the first line of $\mathcal{L}_{AE}$ is the reconstruction loss, which scores the encoder quality and optimizes binary codes to preserve the local structure of the input feature space. The second line is the discriminating loss to encourage distributions of generated variables $\mathbf{b}$ and $\mathbf{z}'$ resemble more the targeted distributions to maximize the entropy. The third line is the regression loss, which provides the direct semantic information to separate codes structure more for different labels. The fourth line is the regularization term of the classifier's weights to simplify the network parameters and avoid overfitting. Overall, $\mathcal{L}_{AE}$ is optimized over the parameter space $\{\theta_{1},\theta_{2},\theta_{3},\theta_{4},\theta_{5}\}$ to train the binary codes of good quality.
	
	\subsection{Out-of-Sample Extension}
	
	After STBH is trained, given an unkown data $\mathbf{x}^{q}$, its binary codes $\mathbf{b}^{q}$ can be generated by the encoder of the network as follows:
	\begin{equation}
		\begin{aligned}
			\mathbf{b}^{q} & = \left( \mathtt{sgn}\left[f_{1}(\mathbf{x}^{q};\theta_{1})-0.5\right] +1\right)/2 \\
			& \in \{0,1\}^{M}
		\end{aligned}
	\end{equation}
	In fact, it is a special case of Eqn. \ref{eqn:neuron} where the random variable $\epsilon^{i}$ is fixed to $0.5$.
	
	\section{Experiments}
	
	We evaludate the performance of STBH model on three widely used image benchmarks: CIFAR-10, NUS-WIDE and MS-COCO such that we can make comparisons with the original TBH and other papers.
	
	\subsection{Implementation Details}
	The model is implemented with Tensorflow \cite{abadi2016tensorflow}, and the code for the TBH network is forked from the original TBH paper \cite{shen2020auto}. A convolutional network called InceptionV3 \cite{DBLP:journals/corr/SzegedyVISW15} is utilized here to extract the features from all datasets. For the experimental parameters, we set the training batch size to 1500, and the length of binary codes $M=32$. In addition, we do a grid search to fine-tune the weights $\gamma$ and $\eta$. Other parameters follow the TBH paper \cite{shen2020auto}. Besides, we perform a six-fold cross validation for all experiments and report the average for the fair comparisons. Our implementation code can be found at \href{https://github.com/Wallace-Chen/TBH/tree/supervised}{https://github.com/Wallace-Chen/TBH/tree/supervised}.
	
	\subsection{Datasets}
	\textbf{CIFAR-10} \cite{krizhevsky2009learning} is a single-label dataset with 60,000 images from 10 classes. We randomly sample 10,000 images as the query set and the remaining 50,000 images as the training set. The dataset analysed during the current study is available at \href{https://www.cs.toronto.edu/~kriz/cifar.html}{https://www.cs.toronto.edu/~kriz/cifar.html}.
	
	\noindent \textbf{NUS-WIDE} \cite{chua2009nus} is a multi-label dataset consisting of 269,648 images from 81 labels. Following the setting in the paper \cite{xia2014supervised}, we select a subset of images from 21 most frequent labels, where the number of images associated with the tag is at least 5,000. From the subset, we randomly choose 10,000 images as the query set and 50,000 images as the training set. The dataset analysed during the current study is available at \href{https://lms.comp.nus.edu.sg/wp-content/uploads/2019/research/nuswide/NUS-WIDE.html}{https://lms.comp.nus.edu.sg/wp-content/uploads/2019/research/nuswide/NUS-WIDE.html}.
	
\begin{figure}[!h]
		\center
		\includegraphics[width=0.9\linewidth]{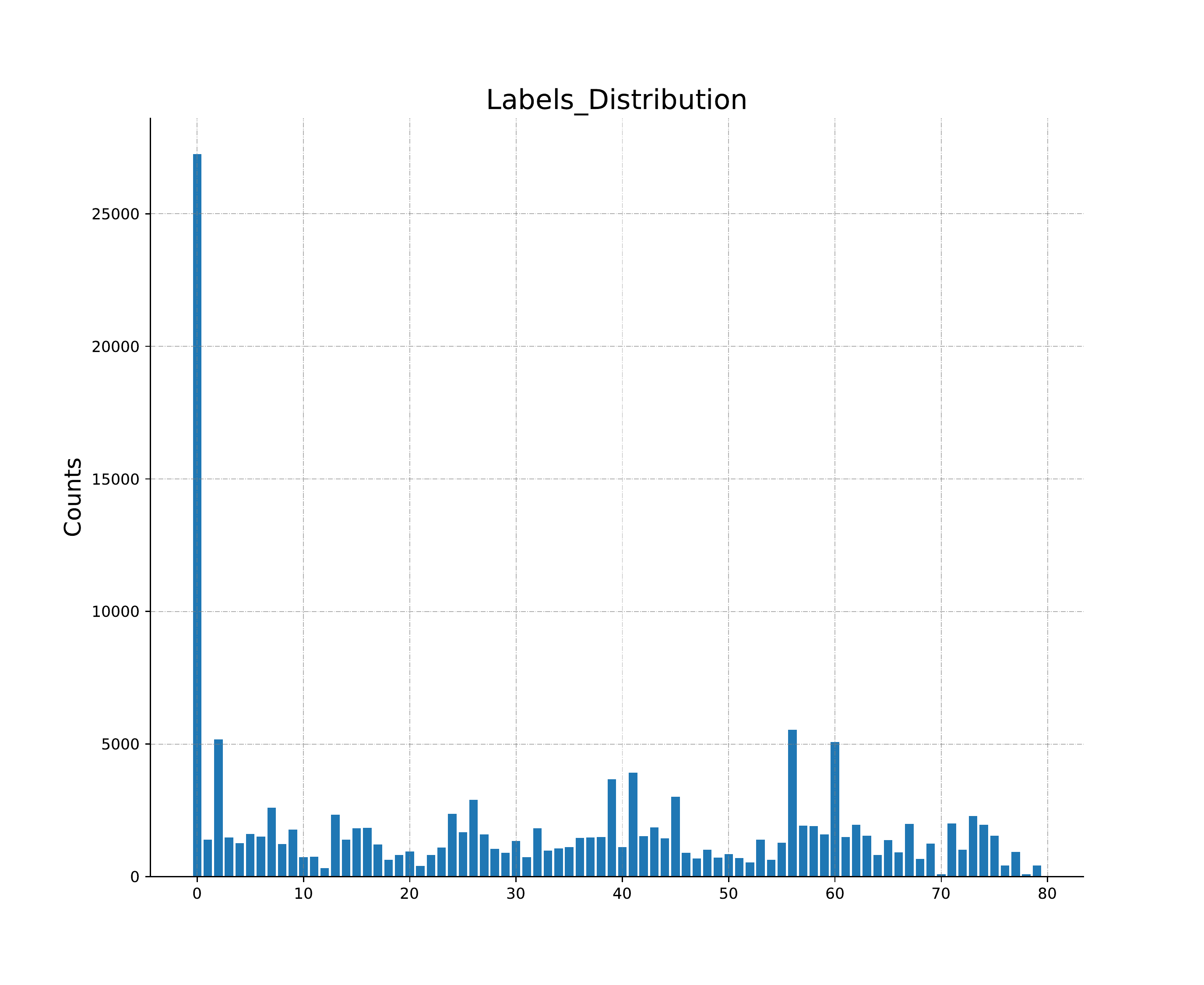}
		\caption{Label distribution of randomly chosen 50,000 images from MS-COCO dataset with 80 classes. More than half of images are associated by the 'person' (first) label.}
		\label{fig:ms-coco_class80_label}
	\end{figure}

	\noindent \textbf{MS-COCO} \cite{lin2014microsoft} is another multi-label dataset. Here, we adopt the 2014 train/val set with 12,2218 images of 80 categories. However, we observe that the "person" label dominates among all classes. For example, as shown in Fig. \ref{fig:ms-coco_class80_label}, more than half of the images are associated with the "person" label. This would be a serious class-imbalance issue when we apply the supervised learning (for more discussions see \ref{sec:class_imbalance}). Since the dataset is multi-labeled, it is not trivial to rebalance different classes. We decided to remove the "person" class and also the images labeled with this class. We randomly pick 8,000 images as the query set and another 40,000 images as the training set from the remaining set. The resulting label distribution is shown in Fig. \ref{fig:label_dist_c}. The dataset analysed during the current study is available at \href{https://cocodataset.org/#download}{https://cocodataset.org/\#download}.
	
	\subsection{Metrics}
	To evaluate the performance of our model, we adopt the standard metrics: Precision w.r.t. top 1000 returned items (P@1000), Precision-Recall curve, and mean Average Precision (mAP). For multi-label datasets (NUS-WIDE and MS-COCO), we regard two items similar when they share at least one common label. In addition, we report the average and the standard deviation to have a fair comparison between the STBH and TBH  models. To make the comparison statistically significant, p-value (see Appendix \ref{sec:appendix_pvalue} for the details) is computed for the STBH against the TBH model.
	
	\subsection{Results Comparison}
	
	\begin{table*}[!h]
		\centering
		\fontsize{9}{12}\selectfont
		\begin{threeparttable}
			\caption{Performance comparison in terms of P@1000 Between the TBH model and STBH model with various $\gamma=\eta$}
			\begin{tabular}{c|c|c|c|c|c|c|c|c|c}
				\toprule
				\multirow {2}{*}{\textbf{Method}}&\multicolumn{3}{|c}{\textbf{CIFAR-10}} &\multicolumn{3}{|c}{\textbf{NUS-WIDE}} &\multicolumn{3}{|c}{\textbf{MS-COCO ("person" label removed)}} \cr
				\cmidrule(lr){2-4} \cmidrule(lr){5-7} \cmidrule(lr){8-10}
				\!&mean\!&s.t.d.\!&p-value\!&mean\!&s.t.d.\!&p-value\!&mean\!&s.t.d.\!&p-value\!\cr
				\midrule
				
				\textbf{TBH} & 0.590 & 0.010 & - & 0.680 & 0.004 & - & 0.670 & 0.008 & - \cr
				\hline 
				\textbf{STBH}, $\gamma=\eta=1$ & 0.600 & 0.010 & 0.4452 & 0.706 & 0.005 & 0.2772 & 0.680 & 0.008 & 0.434 \cr
				\textbf{STBH}, $\gamma=\eta=10$ & 0.710 & 0.008 & 0.0468 & 0.789 & 0.002 & 0.0050 & 0.703 & 0.011 & 0.308 \cr
				\textbf{STBH}, $\gamma=\eta=20$ & 0.796 & 0.004 & 0.0031 & 0.808 & 0.004 & 0.0031 &0.727 & 0.028 & 0.266 \cr
				\textbf{STBH}, $\gamma=\eta=30$ & 0.832 & 0.005 & 0.0012 & 0.826 & 0.004 & 0.0013 & \textbf{0.769} & \textbf{0.009} & \textbf{0.064} \cr
				\textbf{STBH}, $\gamma=\eta=40$ & 0.852 & 0.002 & 0.0008 & 0.838 & 0.004 & 0.0008 & 0.765 & 0.013 & 0.092 \cr
				\textbf{STBH}, $\gamma=\eta=50$ & \textbf{0.861} & \textbf{0.003} & \textbf{0.0006} & \textbf{0.845} & \textbf{0.003} & \textbf{0.0003} & 0.764 & 0.009 & 0.077 \cr
				\bottomrule
			\end{tabular}\label{tab:precision}
		\end{threeparttable}
	\end{table*}
	
	\begin{table*}[!h]
		\centering
		\fontsize{9}{12}\selectfont
		\begin{threeparttable}
			\caption{Performance comparison in terms of mAP Between the TBH model and STBH model with various $\gamma=\eta$}
			\begin{tabular}{c|c|c|c|c|c|c|c|c|c}
				\toprule
				\multirow {2}{*}{\textbf{Method}}&\multicolumn{3}{|c}{\textbf{CIFAR-10}} &\multicolumn{3}{|c}{\textbf{NUS-WIDE}} &\multicolumn{3}{|c}{\textbf{MS-COCO ("person" label removed)}} \cr
				\cmidrule(lr){2-4} \cmidrule(lr){5-7} \cmidrule(lr){8-10}
				\!&mean\!&s.t.d.\!&p-value\!&mean\!&s.t.d.\!&p-value\!&mean\!&s.t.d.\!&p-value\!\cr
				\midrule
				
				\textbf{TBH} & 0.636 & 0.009 & - & 0.703 & 0.004 & - & 0.570 & 0.010 & - \cr
				\hline 
				\textbf{STBH}, $\gamma=\eta=1$ & 0.650 & 0.008 & 0.4216 & 0.733 & 0.005 & 0.2447 & 0.580 & 0.009 & 0.447 \cr
				\textbf{STBH}, $\gamma=\eta=10$ & 0.750 & 0.007 & 0.0440 & 0.811 & 0.002 & 0.0052 & 0.630 & 0.012 & 0.222 \cr
				\textbf{STBH}, $\gamma=\eta=20$ & 0.824 & 0.003 & 0.0034 & 0.827 & 0.003 & 0.0014 &0.653 & 0.025 & 0.198 \cr
				\textbf{STBH}, $\gamma=\eta=30$ & 0.851 & 0.005 & 0.0017 & 0.842 & 0.003 & 0.0014 & \textbf{0.696} & \textbf{0.006} & \textbf{0.038} \cr
				\textbf{STBH}, $\gamma=\eta=40$ & 0.865 & 0.002 & 0.0011 & 0.853 & 0.004 & 0.0011 & 0.695 & 0.014 & 0.064 \cr
				\textbf{STBH}, $\gamma=\eta=50$ & \textbf{0.873} & \textbf{0.003} & \textbf{0.0009} & \textbf{0.859} & \textbf{0.002} & \textbf{0.0005} & 0.695 & 0.009 & 0.049 \cr
				\bottomrule
			\end{tabular}\label{tab:map}
		\end{threeparttable}
	\end{table*}
	
	\begin{figure*}[!h]
		\centering 
		\subfigure[P-R Curve on CIFAR-10]{
			\includegraphics[width=0.3\textwidth]{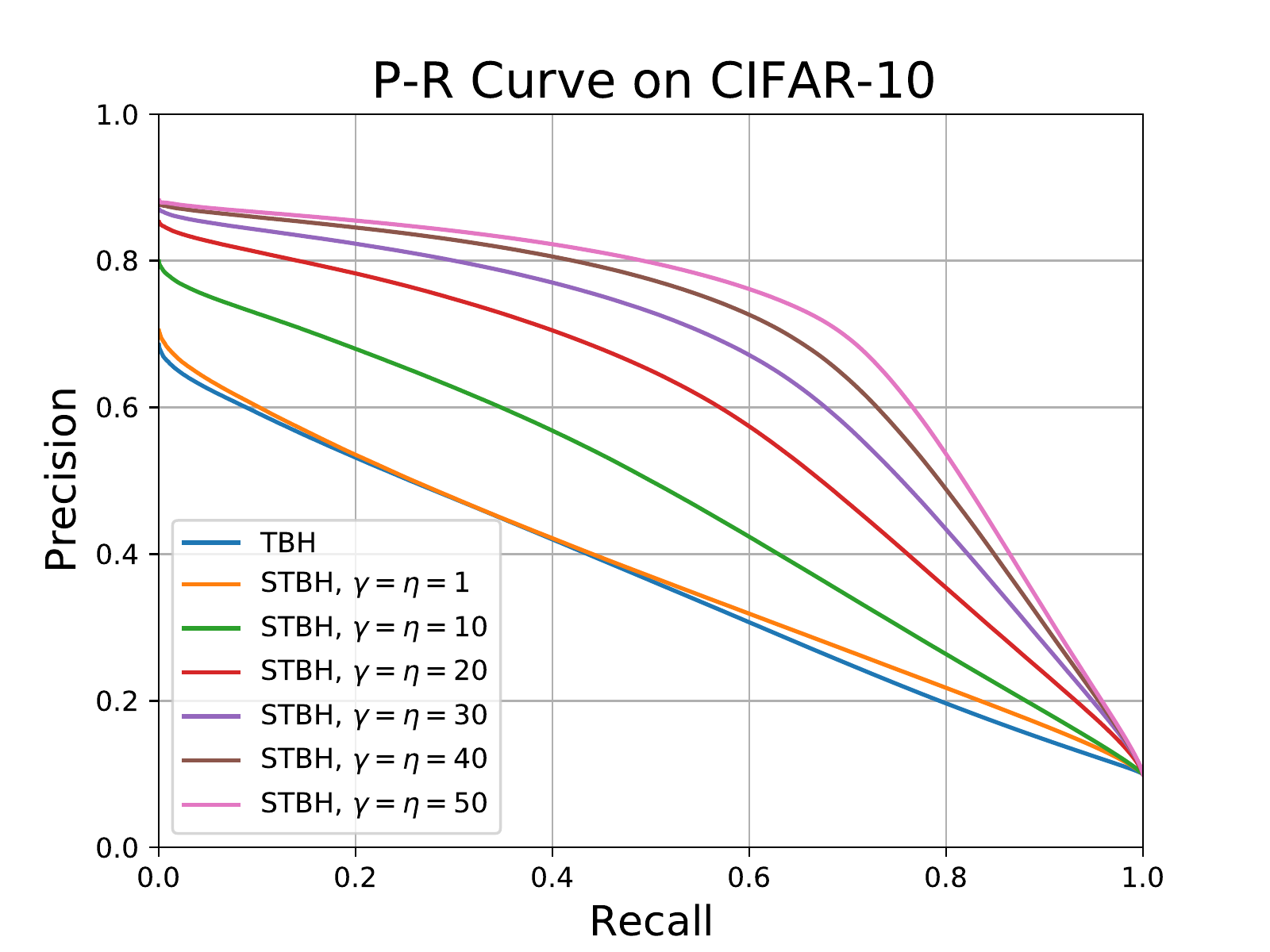}}
		\subfigure[P-R Curve on NUS-WIDE]{
			\includegraphics[width=0.3\textwidth]{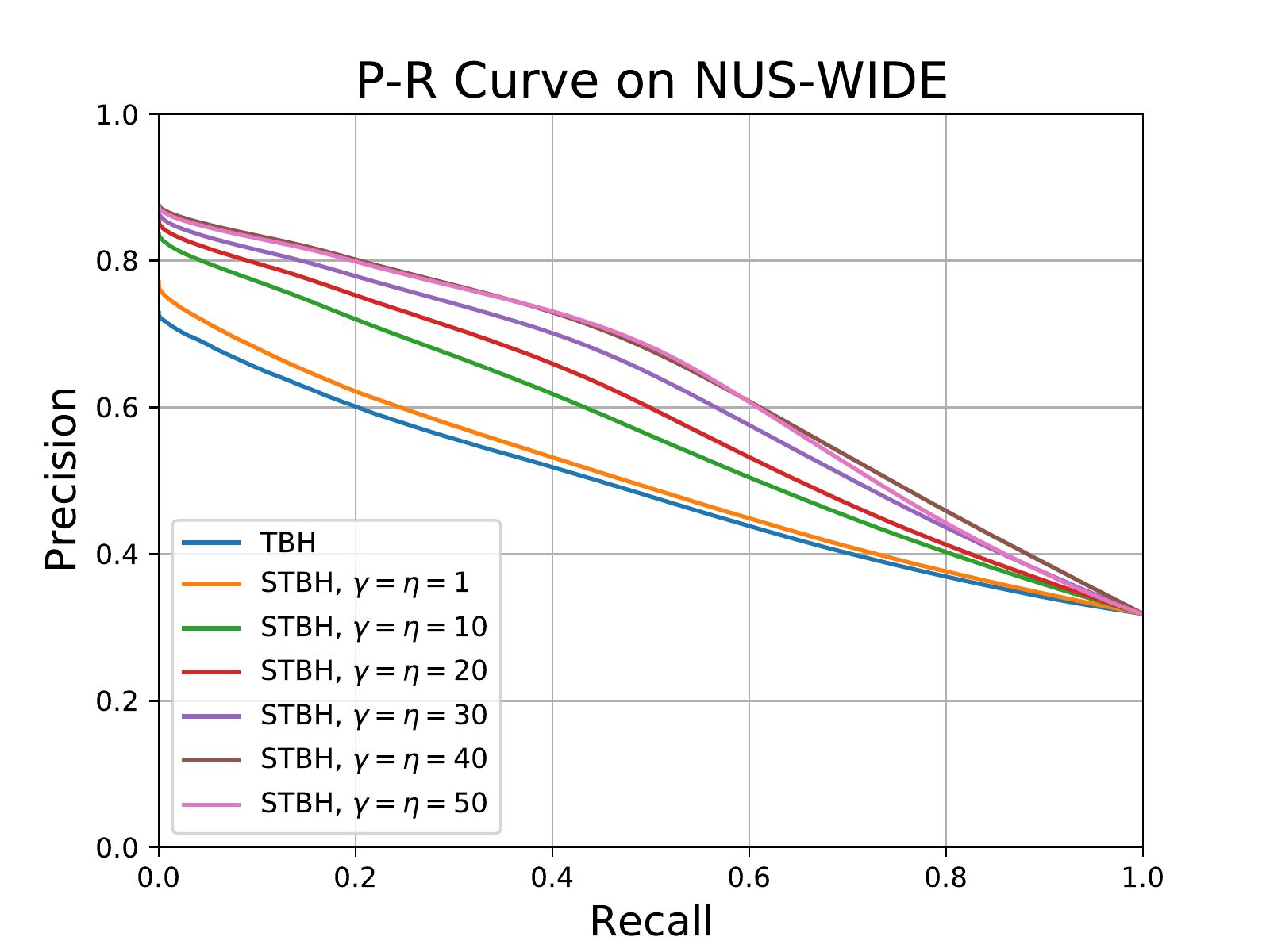}}
		\subfigure[P-R Curve on MS-COCO]{
			\includegraphics[width=0.3\textwidth]{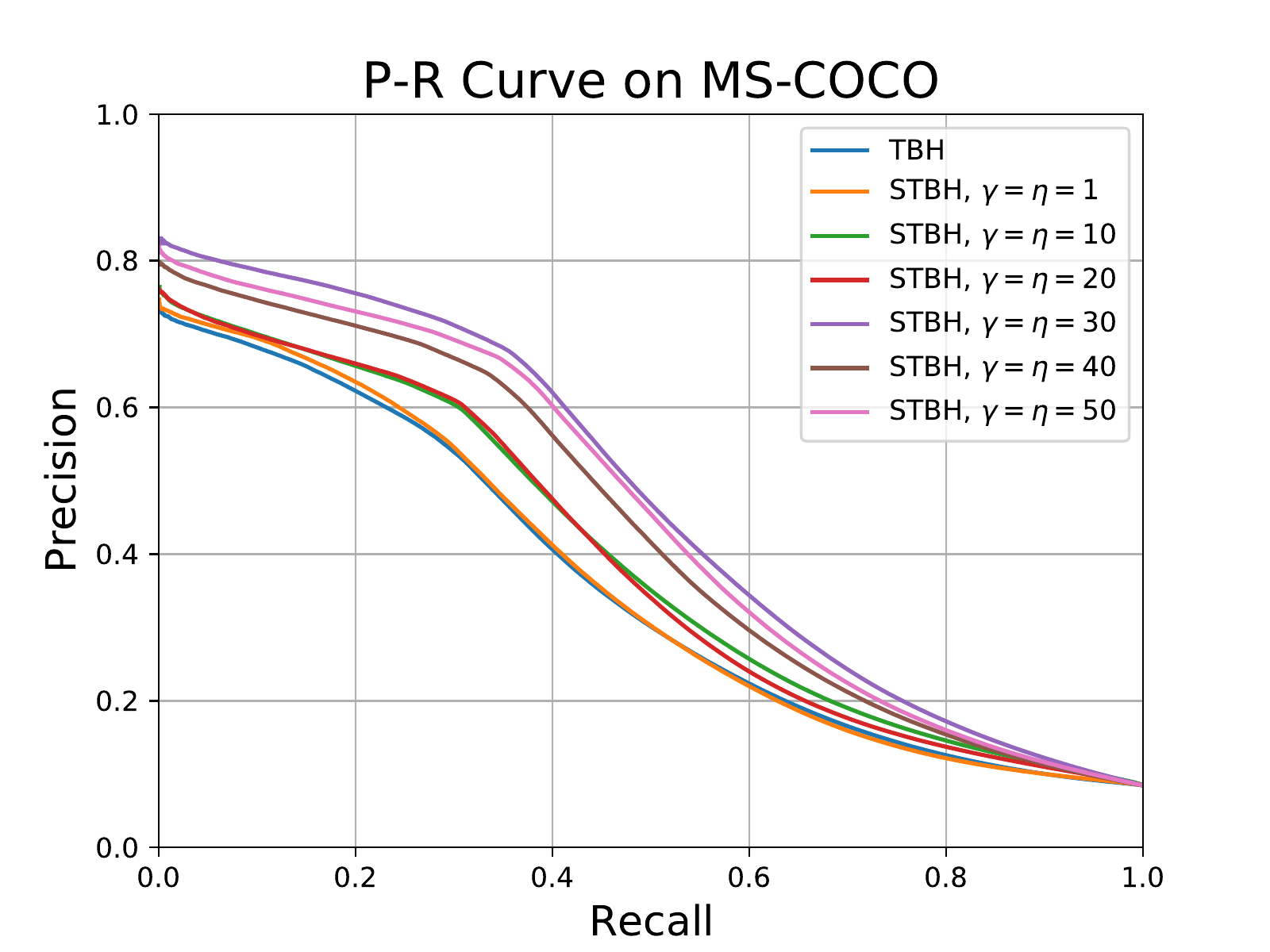}}
		\caption{Comparisons of P-R Curves for the CIFAR-10 (left), NUS-WIDE (midlle), and MS-COCO (right) }
		\label{fig:pr_curve}
	\end{figure*}
	
	\begin{figure*}[!h]
		\centering 
		\subfigure[CIFAR-10 (50,000 images)]{
			\includegraphics[width=0.3\textwidth]{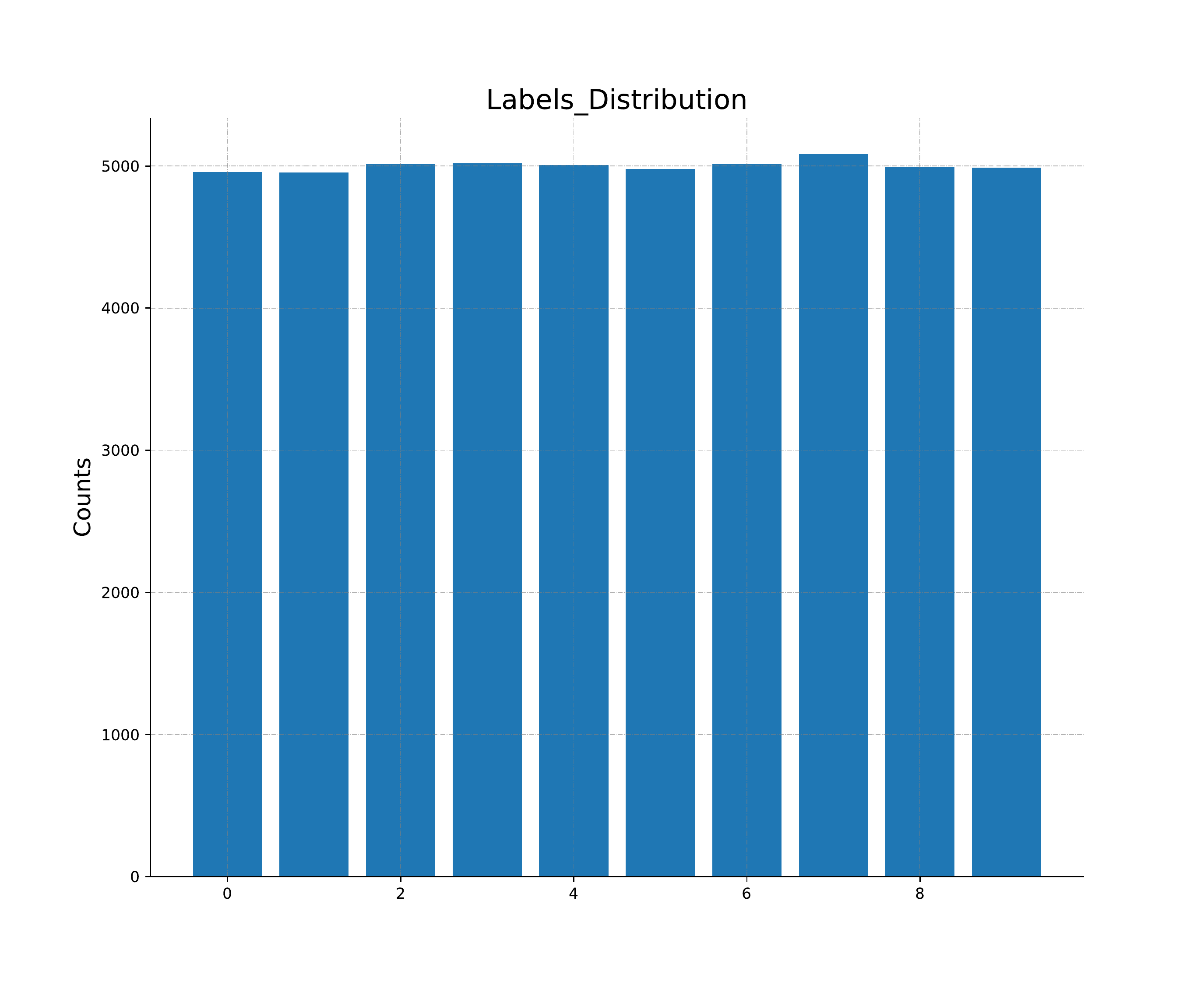}}
		\subfigure[NUS-WIDE (50,000 images)]{
			\includegraphics[width=0.3\textwidth]{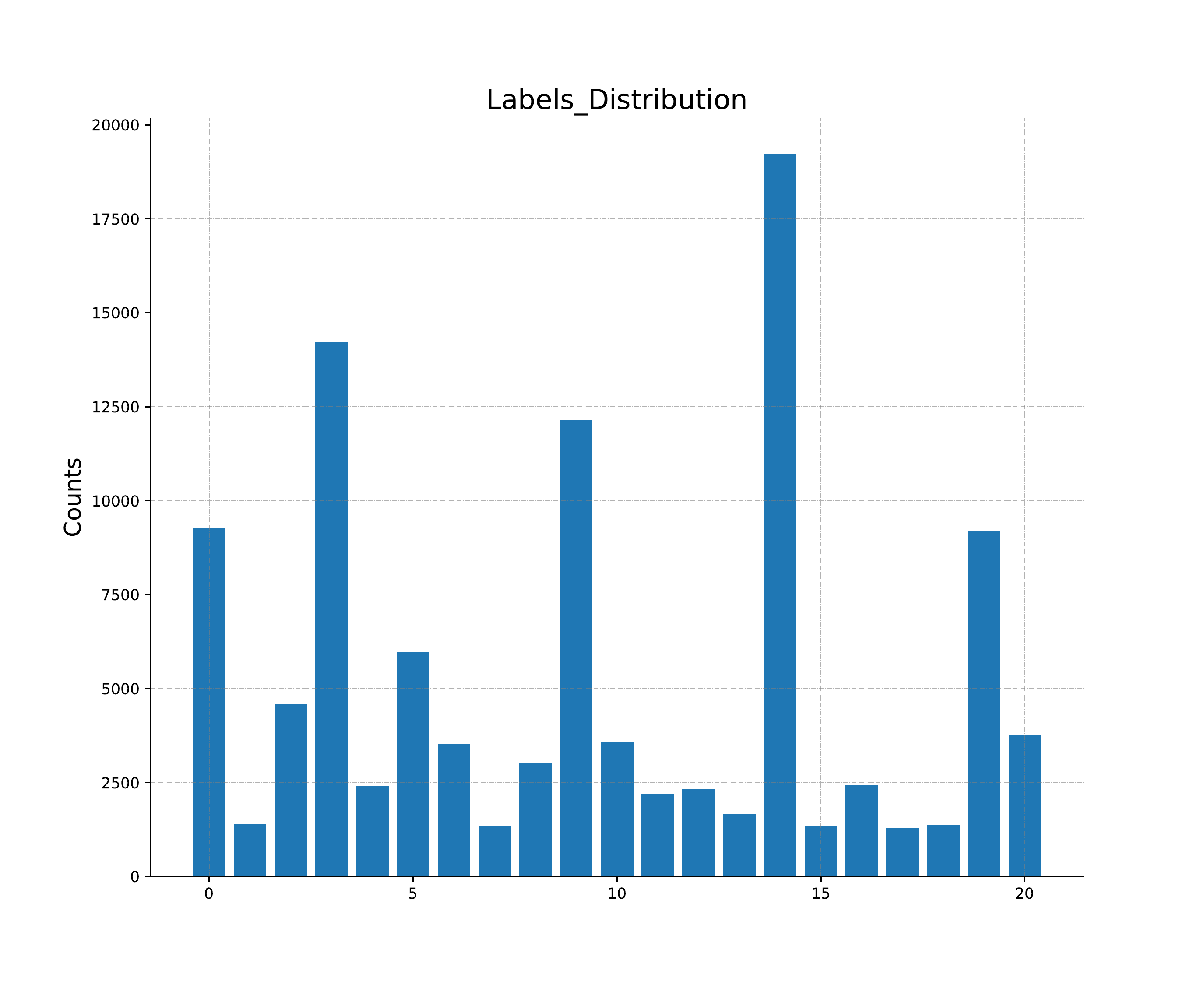}}
		\subfigure[MS-COCO (40,000 images)]{\label{fig:label_dist_c}
			\includegraphics[width=0.3\textwidth]{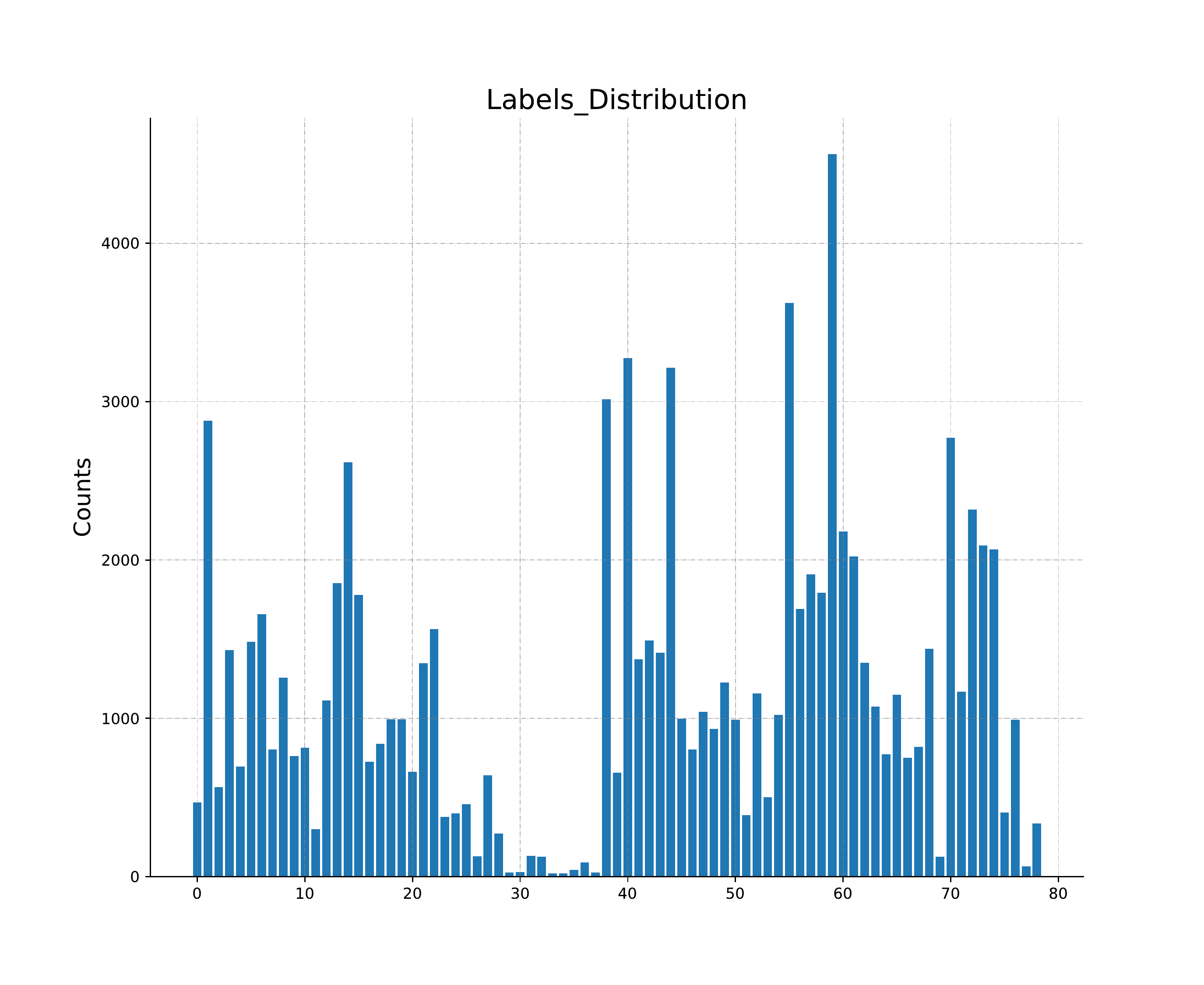}}
		\caption{Class distributions for the CIFAR-10 (left), NUS-WIDE (midlle), and MS-COCO (right). X-axis indicates the class ID while the Y-axis is the counts. CIFAR-10 has the most balanced class distribution. }
		\label{fig:label_dist}
	\end{figure*}
	
	To compare our STBH model to the original TBH model, we keep every condition the same except turning on the regression and regularization term in the loss function for STBH. In addition, a grid search on the value of $\gamma=\eta$ is performed for each dataset.
	
	The results of P@1000 and mAP are reported in Table \ref{tab:precision} and \ref{tab:map} respectively, while the P-R curves are shown in Fig. \ref{fig:pr_curve}. The performance of our proposed STBH gets much improved compared to the original TBH model when $\gamma$ and $\eta$ are carefully chosen. For example, setting $\gamma=\eta=20$ for the CIFAR-10 and NUS-WIDE, and $\gamma=\eta=30$ for the MS-COCO are good options to achieve a remarkable improvement. From the point of view of p-value, if we set the significance level to $0.05$, the above numbers can also give us p-values below or around the significance level, which means results of STBH are better than those of TBH at the confidence level of $95\%$. The great improvements demonstrate the view that semantic supervision is very helpful and powerful for solving the hashing problem and can provide extra information other than that contained in the features extracted from the raw dataset.
	
	\subsection{Class Imbalance}
	\label{sec:class_imbalance}
	
	Another observation from the above results is that the performance improvement of the CIFAR-10 dataset is the greatest among the three datasets because we have a more balanced class distribution in CIFAR-10, as shown in Fig. \ref{fig:label_dist}. The class imbalance in the other two multi-label datasets NUS-WIDE and MS-COCO affects the ability of supervised learning.
	
	The uneven class distribution has been a challenge for the classification problem, as the design of most classifiers aims to reduce the global error, which tends to benefit and gets biased towards the majority class, leading to the poor accuracy on the minority examples and underperformance  \cite{krawczyk2016learning}\cite{fernandez2018smote}\cite{charte2015mlsmote}. There exist many works that try to solve the problem in the multi-label datasets, which can be categorized into three approaches \cite{charte2015mlsmote}: algorithmic adaptations, ensemble-based methods and resampling techniques. Algorithmic adaptations \cite{he2012imbalanced}\cite{li2013improvement}\cite{tepvorachai2008multi}\cite{chen2006efficient} aim to design classifiers insensitive to the imbalanced datasets and are mostly specific to the domains considered in the problem. Ensemble-based methods \cite{tahir2012multilabel}\cite{tahir2012inverse} normally build ensembles of classifiers and then obtain a final prediction. However they require the training of many classifiers and suffer from low efficiency. Resampling techniques \cite{igarashi20103d}\cite{charte2014jesus}\cite{giraldo2013managing} instead focus on preprocessing of the dataset to produce a more balanced version, and involve undersampling or/and oversampling. MLSMOTE \cite{charte2015mlsmote}, for example, is one of extensions of original SMOTE \cite{chawla2002smote} method designed for multi-label datasets.
	
	For the datasets used in this paper, NUS-WIDE and MS-COCO are multi-label datasets and have skewed class distributions. Especially the imbalance ratio of MS-COCO is quite extreme (1:1000 for some classes) even after the dominated label "person" is removed. A more careful treatment of such imbalance problems is needed for a better performance and left for further work.
	
	
	\subsection{Complexity}
	
	The benefit brought by incorporating label information is remarkable, while the change in the network architecture is minimal, as indicated in Fig. \ref{fig:stbh}. In this section, we show in fact the overhead of running the STBH model is also negligible or small compared to the TBH model in terms of training time, total loss and the memory used.
	
\begin{figure*}[!h]
		\centering 
		\subfigure[CIFAR-10, running time]{
			\label{fig:complexity_a}
			\includegraphics[width=0.32\textwidth]{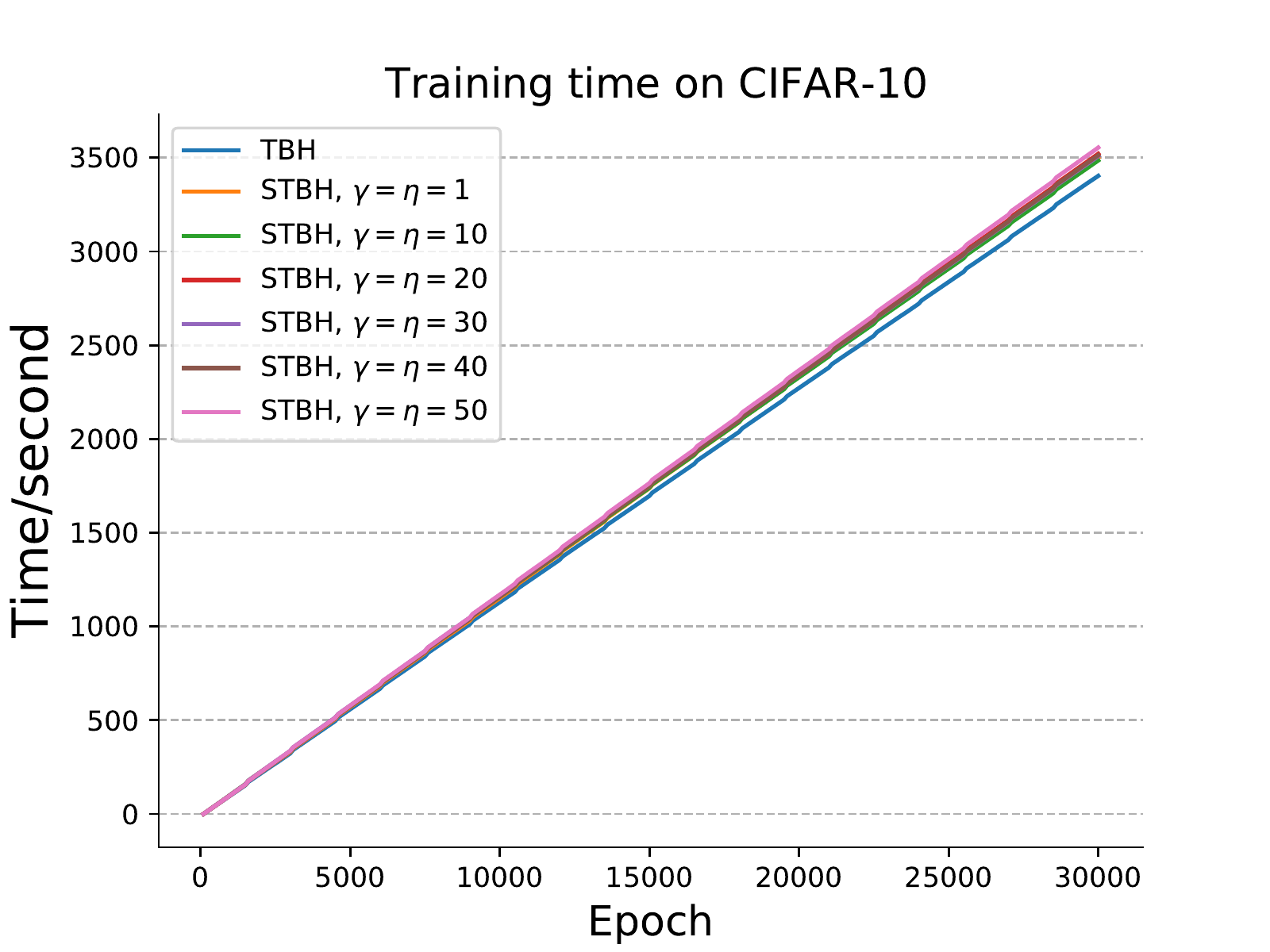}}\hfill
		\subfigure[NUS-WIDE, running time]{
			\label{fig:complexity_b}
			\includegraphics[width=0.32\textwidth]{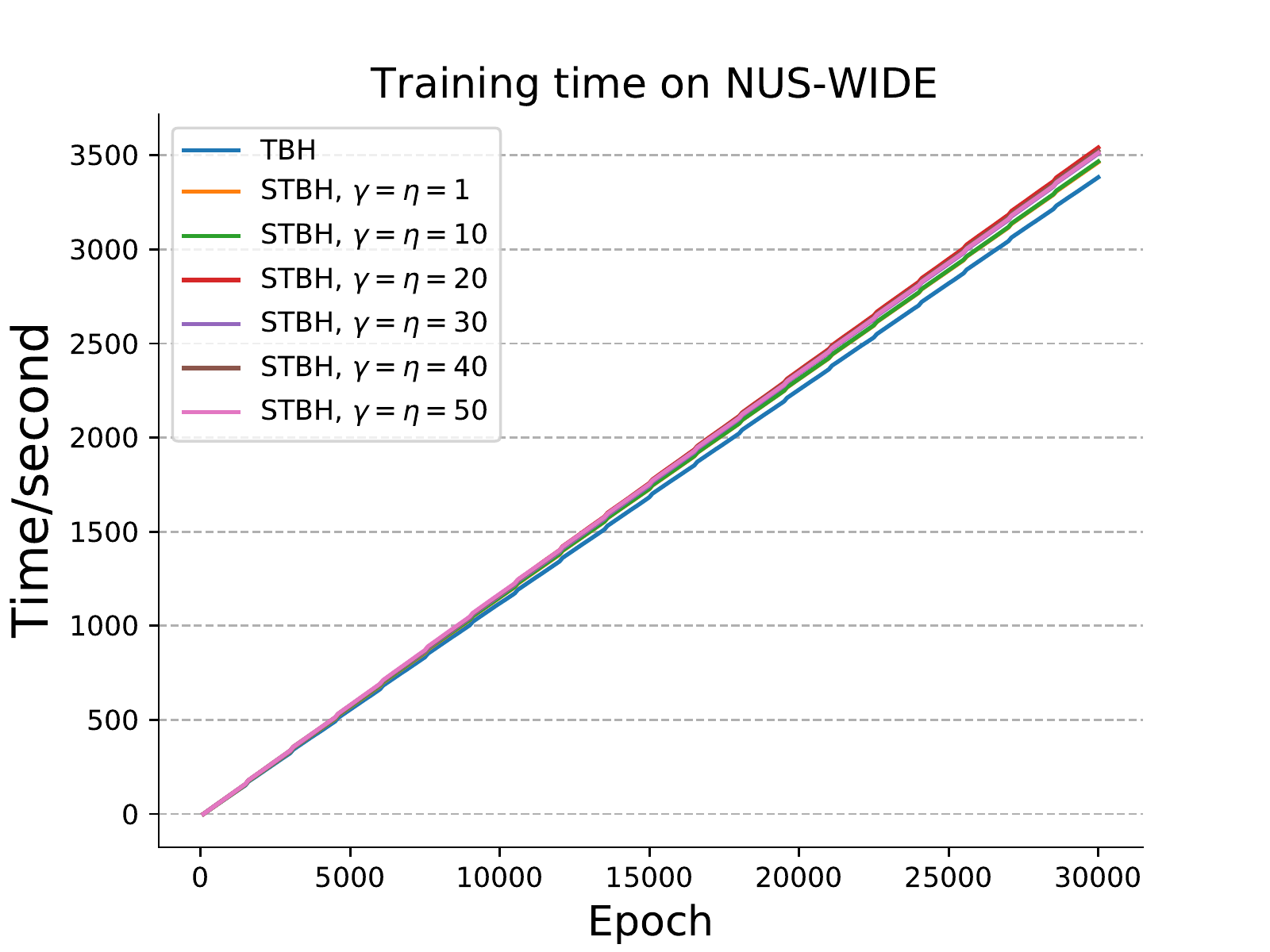}}\hfill
		\subfigure[MS-COCO, running time]{
			\label{fig:complexity_c}
			\includegraphics[width=0.32\textwidth]{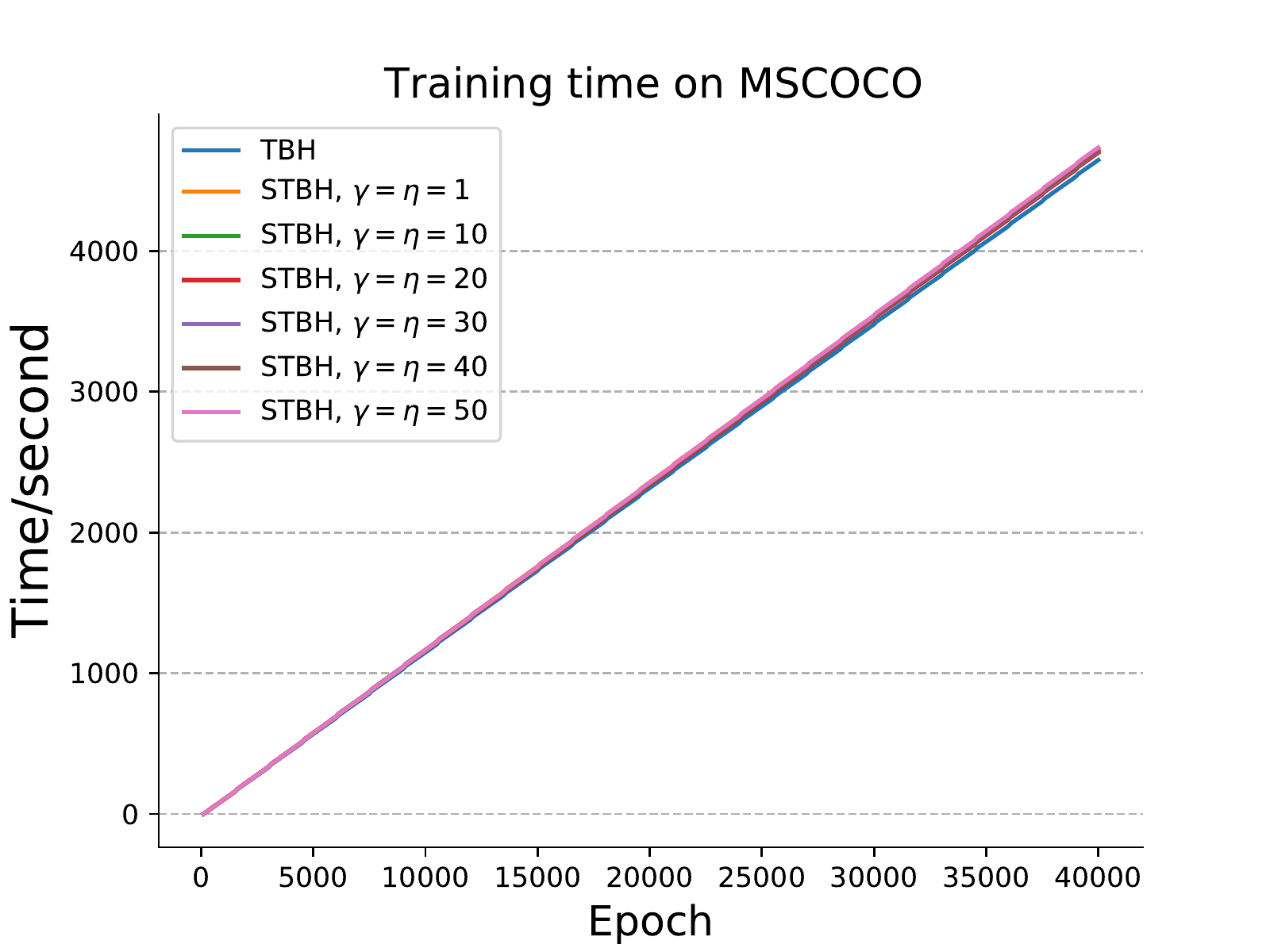}}\\
		\subfigure[CIFAR-10, training loss]{
			\label{fig:complexity_d}
			\includegraphics[width=0.32\textwidth]{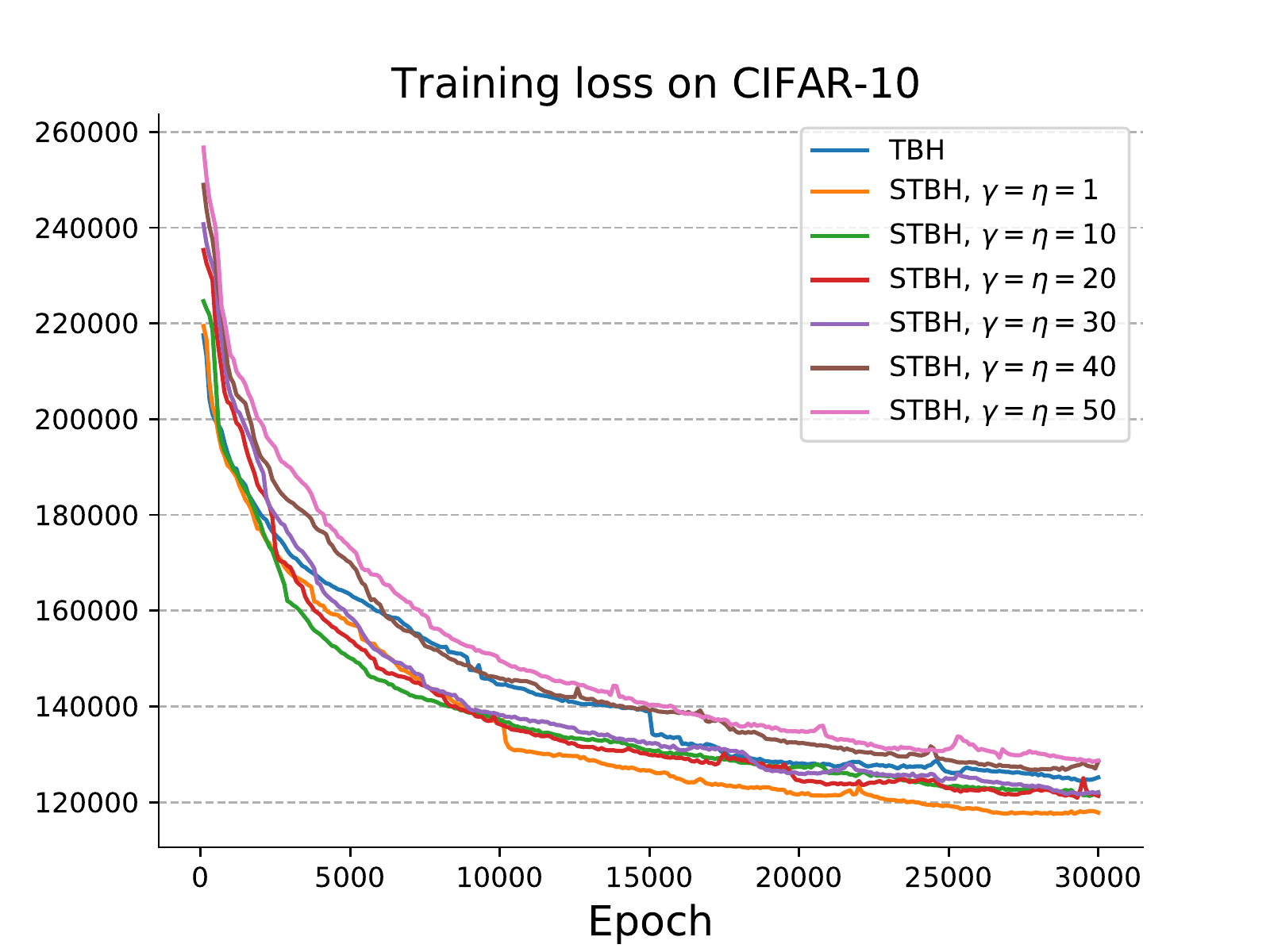}}\hfill
		\subfigure[NUS-WIDE, training loss]{
			\label{fig:complexity_e}
			\includegraphics[width=0.32\textwidth]{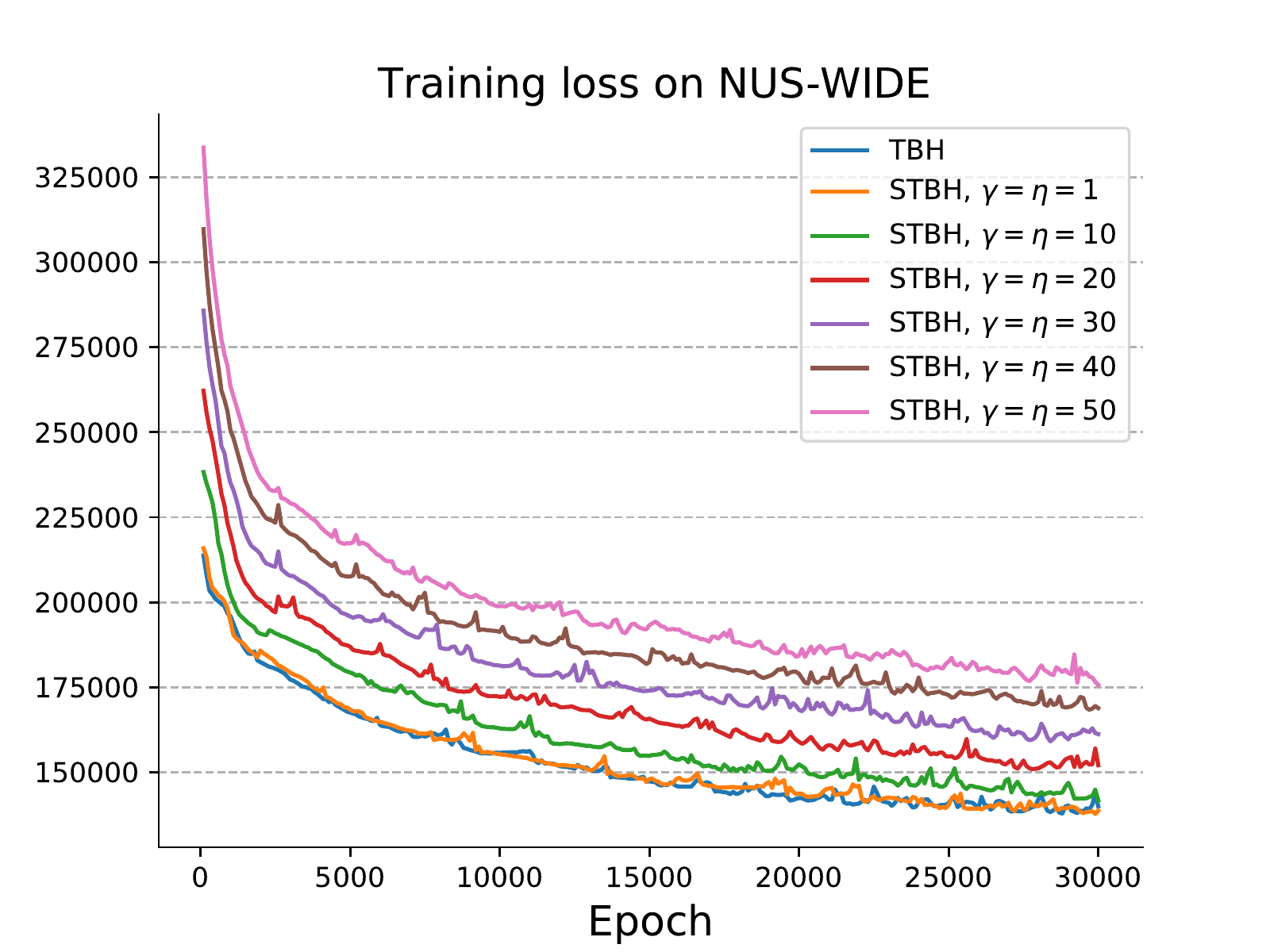}}\hfill
		\subfigure[MS-COCO, training loss]{
			\label{fig:complexity_f}
			\includegraphics[width=0.32\textwidth]{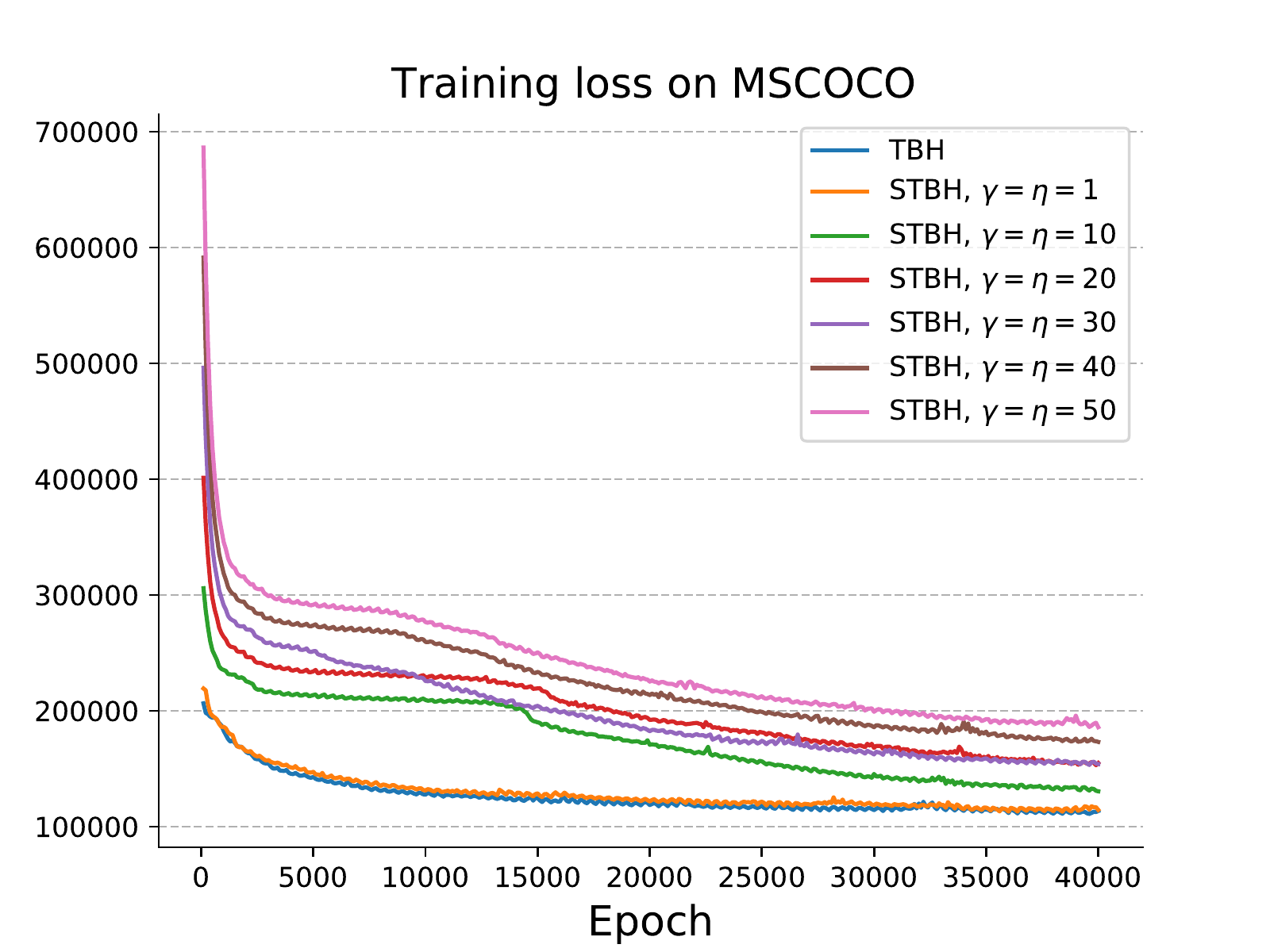}}
		\caption{ Comparisons of training times (upper) and losses (lower) between TBH and STBH with various values of $\gamma(\eta)$ on three datasets.}
		\label{fig:complexity}
	\end{figure*}

	In Fig. \ref{fig:complexity}, the upper plots show the running time over the training steps between TBH and STBH on three datasets, which demonstrates that the STBH model takes the almost same time as the TBH model independent of datasets. The computing platform used is equipped with a 3.6GHz Intel i3-8100 CPU, 32 GB RAM, and NVIDIA RTX 2070.
	
	The lower plots in Fig. \ref{fig:complexity} give the evolution of total training loss between TBH and STBH. As can be clearly seen, the losses are converged at the end of the training, especially compared to those at the start. In addition, STBH can be optimized within the same training steps and does not need extra epochs. Moreover, a closer observation on the subfigure \ref{fig:complexity_d} reveals that STBH with small values of $\gamma(\eta)$ (less than 30 for example) can actually achieve even lower losses than the TBH model. This validates the view that supervision of class labels can provide complementary information and help guide the optimization to find a better solution in the search space.
	
	We also point out here that STBH requires little extra memory space during the training, which can be illustrated by the number of network parameters to be trained. The original TBH model contains $5,543,490$ parameters regardless of datasets, while STBH only needs a few hundreds (CIFAR-10, NUS-WIDE) or thousands (MS-COCO) more parameters, which are negligible. The exact numbers can be found in Table \ref{tab:num_paras}.
	
	It is worthwhile to emphasize that STBH can achieve considerable improvements at virtually little extra cost. 
	
	\begin{table*}[!h]
		\centering
		\fontsize{9}{12}\selectfont
		\caption{Number of network parameters to be trained.}
		\begin{tabular}{c|c|c|c|c}
			\toprule
			\textbf{ } & \textbf{TBH} & \textbf{STBH + CIFAR-10} & \textbf{STBH + NUS-WIDE} & \textbf{STBH + MS-COCO} \cr
			\midrule
			
			\textbf{Number of parameters} & 5,543,490 & 5,543,820 & 5,544,183 & 5,546,097 \cr
			\bottomrule
		\end{tabular}\label{tab:num_paras}
	\end{table*}
	
	\subsection{Grid Search on $\eta$}
	
\begin{figure}[!h]
		\centering 
		\subfigure[Regularization term: $\gamma=\eta=20$]{
			\label{fig:eta_search_cifar10_a}
			\includegraphics[width=0.48\linewidth]{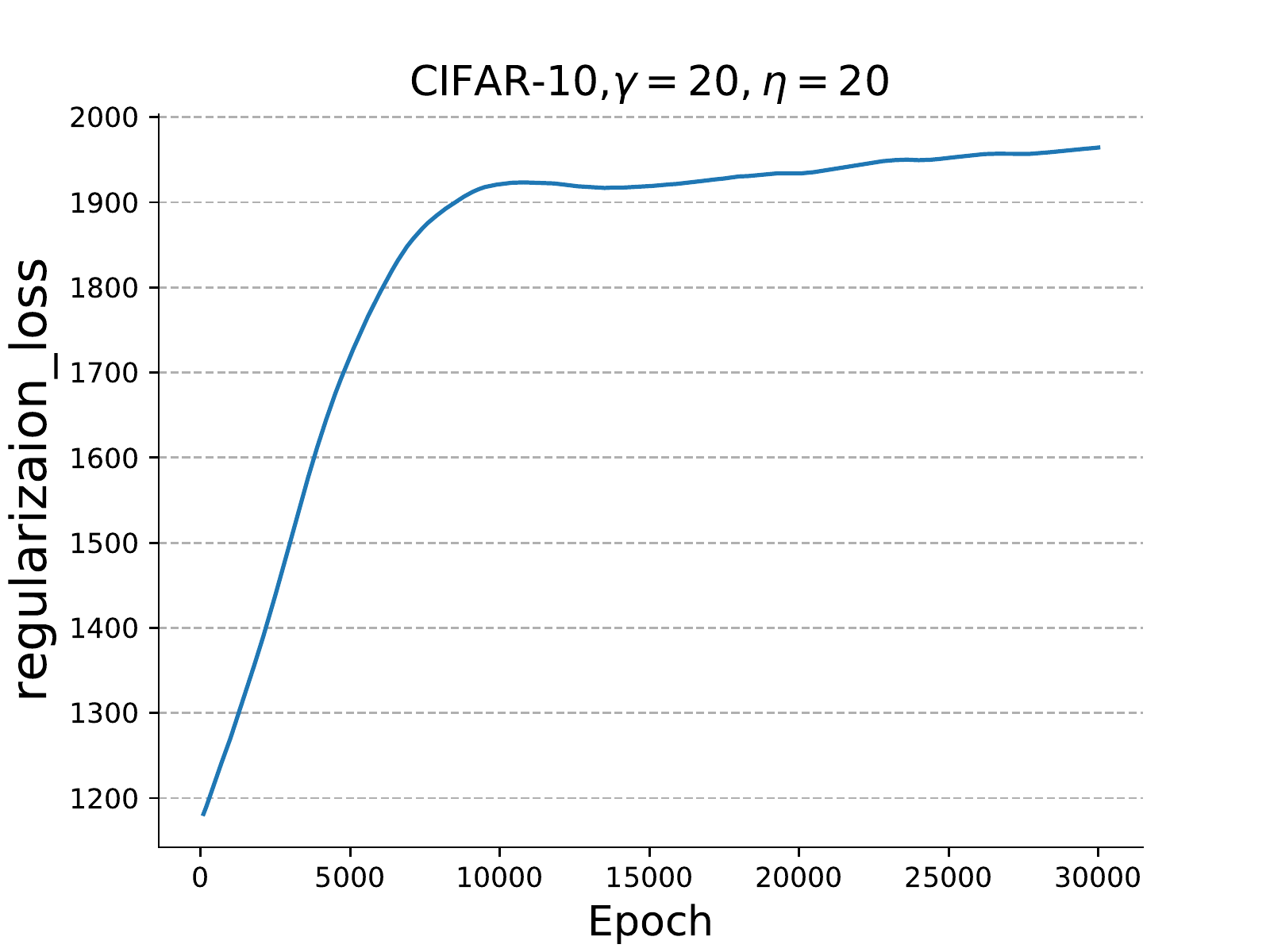}}\hfill
		\subfigure[Regularization term: $\gamma=20,\eta=100$]{
			\label{fig:eta_search_cifar10_b}
			\includegraphics[width=0.48\linewidth]{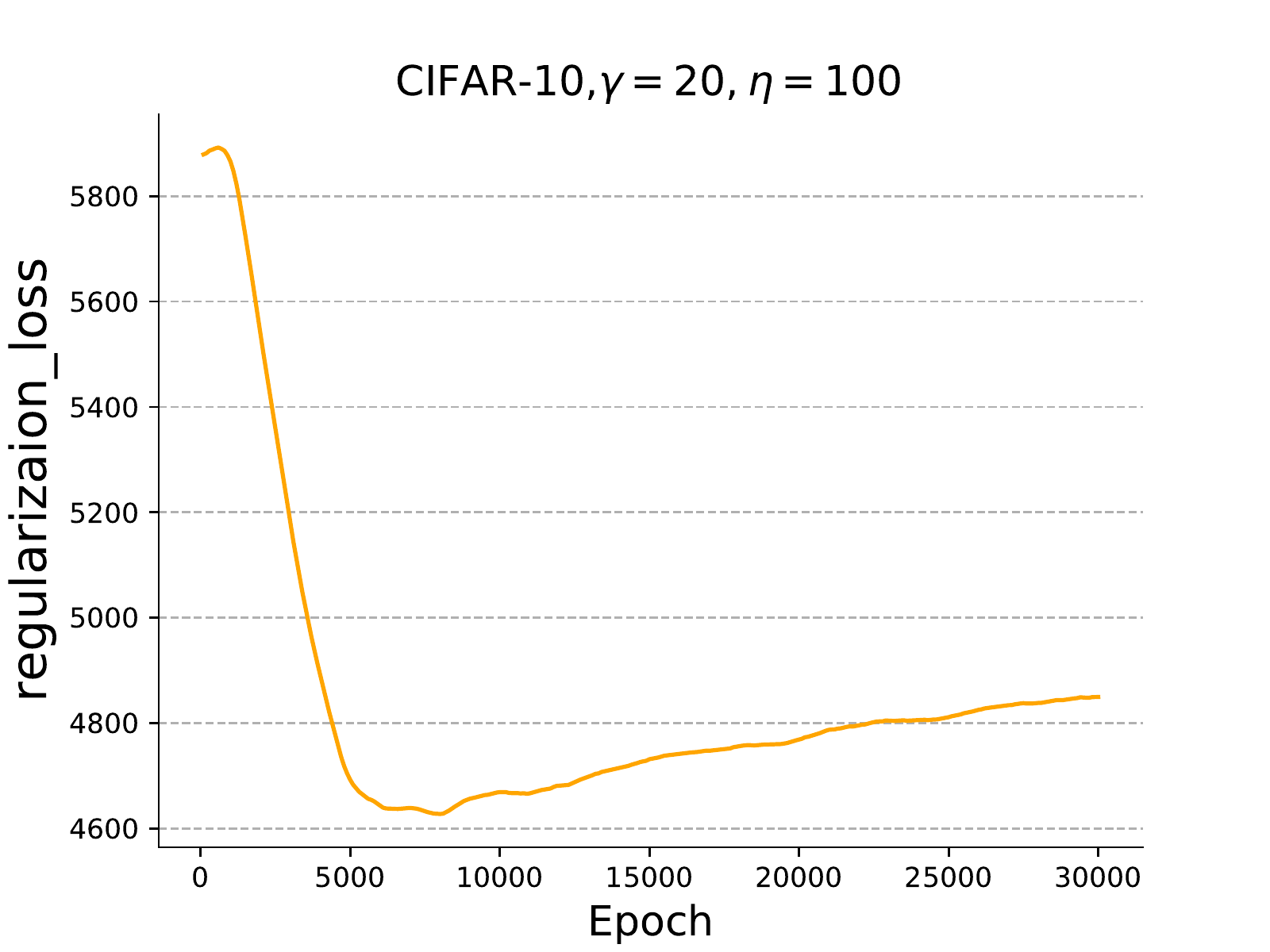}}\\
		\subfigure[Classifier's weights: $\gamma=\eta=20$]{
			\label{fig:eta_search_cifar10_c}
			\includegraphics[width=0.4\linewidth]{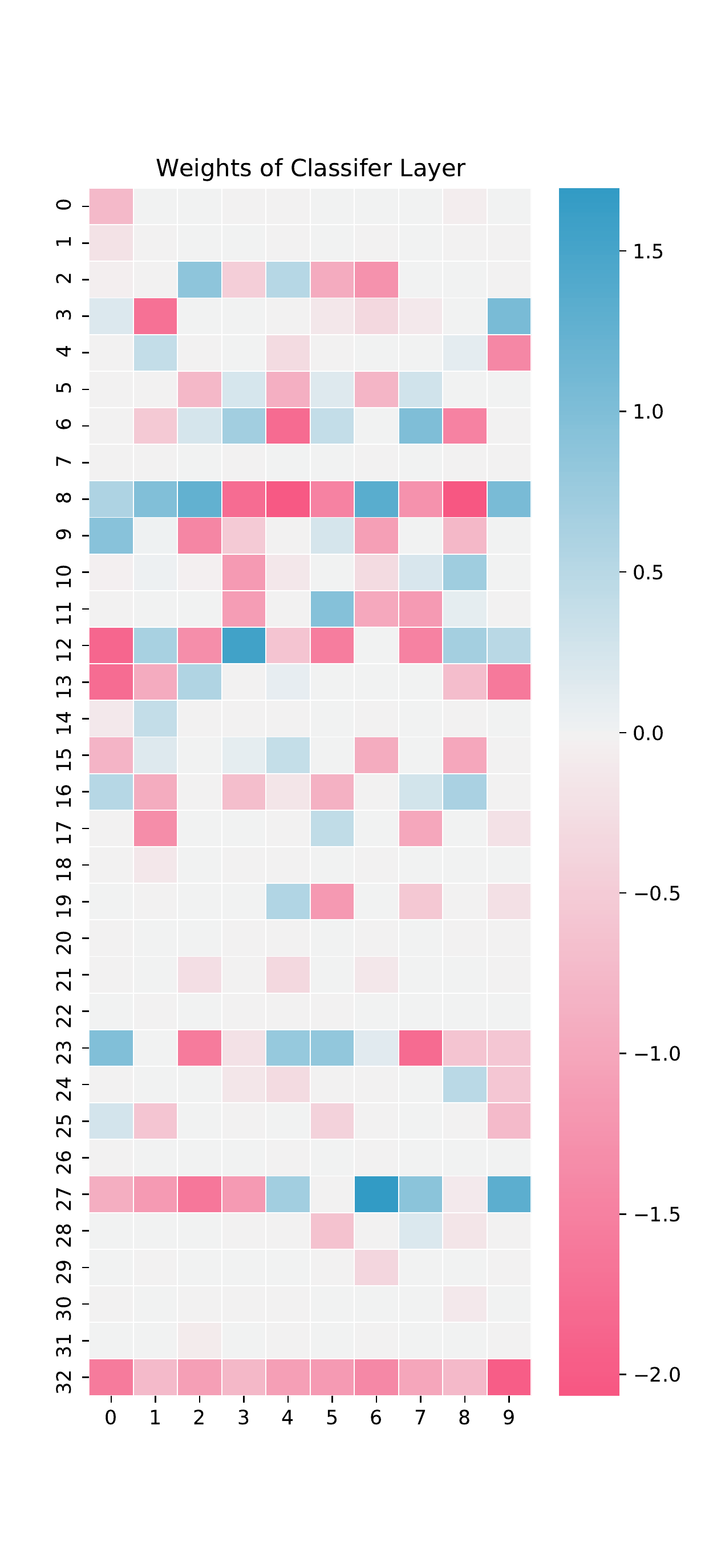}}\hfill
		\subfigure[Classifier's weights: $\gamma=20,\eta=100$]{
			\label{fig:eta_search_cifar10_d}
			\includegraphics[width=0.4\linewidth]{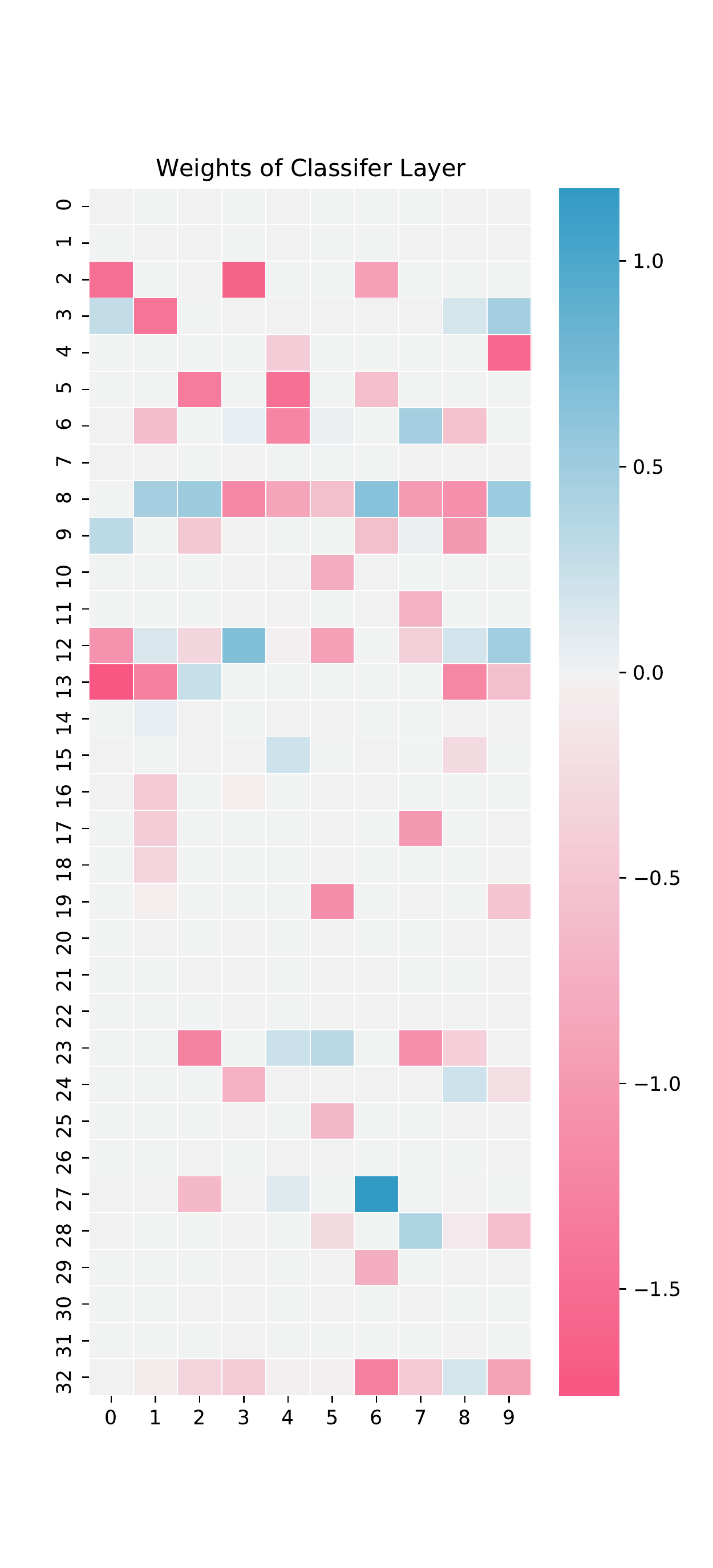}}
		\caption{ Regularization term (upper) and the classifier's weights (lower) for $\eta=20$ (left) and $\eta=100$ (right) while fixing $\gamma=20$, CIFAR-10 dataset. }
		\label{fig:eta_search_cifar10}
	\end{figure}

	We keep $\eta=\gamma$ during the above grid searches, however we could also fine tune the value of $\eta$ while fixing the $\gamma$. Since $\eta$ is the weight of the regularization term of the classifier's weights, we aim to find a proper number of $\eta$ such that the regularization term could get decreased to avoid the overfitting. As shown in Fig. \ref{fig:eta_search_cifar10}, for the CIFAR-10 dataset, the subplots \ref{fig:eta_search_cifar10_a} and \ref{fig:eta_search_cifar10_b} are regularization terms for different $\eta$ values with $\gamma=20$. We can decrease the regularization term by varying the values of $\eta$, and the subplots \ref{fig:eta_search_cifar10_c} and \ref{fig:eta_search_cifar10_d} are the heat maps showing the classifer's weights $\mathbf{W}_{\theta_{5}}$ for the corresponding $\eta$ values respectively. By setting $\eta=100$ instead of $20$, the L1 norm of $\mathbf{W}_{\theta_{5}}$ get much decreased and the parameters are simplified. The price is that the performance of $\eta=100$ gets lowered a bit, but still close to that of $\eta=20$, as can be observed in Table \ref{tab:comp_eta_search_cifar10}.
	
	\begin{table*}[!h]
		\centering
		\fontsize{9}{12}\selectfont
		\caption{Performance comparison on CIFAR-10 with different $\eta$ values.}
		\begin{tabular}{c|c|c|c|c|c|c}
			\toprule
			\multirow {2}{*}{\textbf{Method}}&\multicolumn{3}{|c}{\textbf{P@1000}} &\multicolumn{3}{|c}{\textbf{mAP}} \cr
			\cmidrule(lr){2-4} \cmidrule(lr){5-7}
			\!&mean\!&std.\!&p-value\!&mean\!&std.\!&p-value\!\cr
			\midrule
			
			\textbf{TBH} & 0.590 & 0.010 & - & 0.636 & 0.009 & - \cr
			\hline 
			\textbf{STBH}, $\gamma=20,\eta=20$ & \textbf{0.824} & \textbf{0.002} & \textbf{0.0034} & \textbf{0.796} & \textbf{0.004} & \textbf{0.0031} \cr
			\textbf{STBH}, $\gamma=20,\eta=100$ & 0.802 & 0.002 & 0.0065 & 0.773 & 0.003 & 0.0055 \cr
			\bottomrule
		\end{tabular}\label{tab:comp_eta_search_cifar10}
	\end{table*}
	
	\subsection{Correlation Matrix of Labels}
\begin{figure*}[!h]
		\centering 
		\subfigure[CIFAR-10, TBH]{
			\label{fig:correlation_labels_a}
			\includegraphics[width=0.32\textwidth]{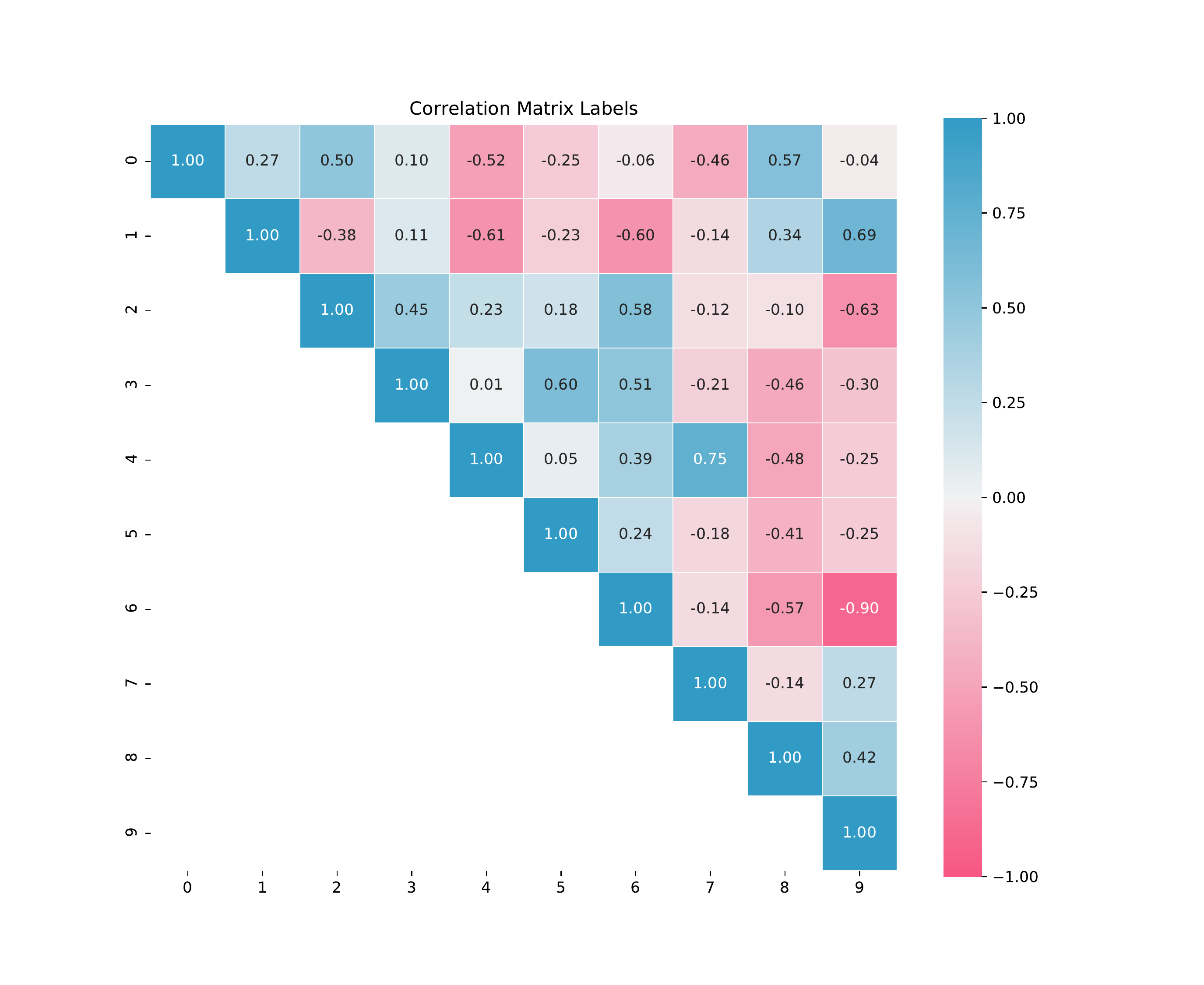}}\hfill
		\subfigure[NUS-WIDE, TBH]{
			\label{fig:correlation_labels_b}
			\includegraphics[width=0.32\textwidth]{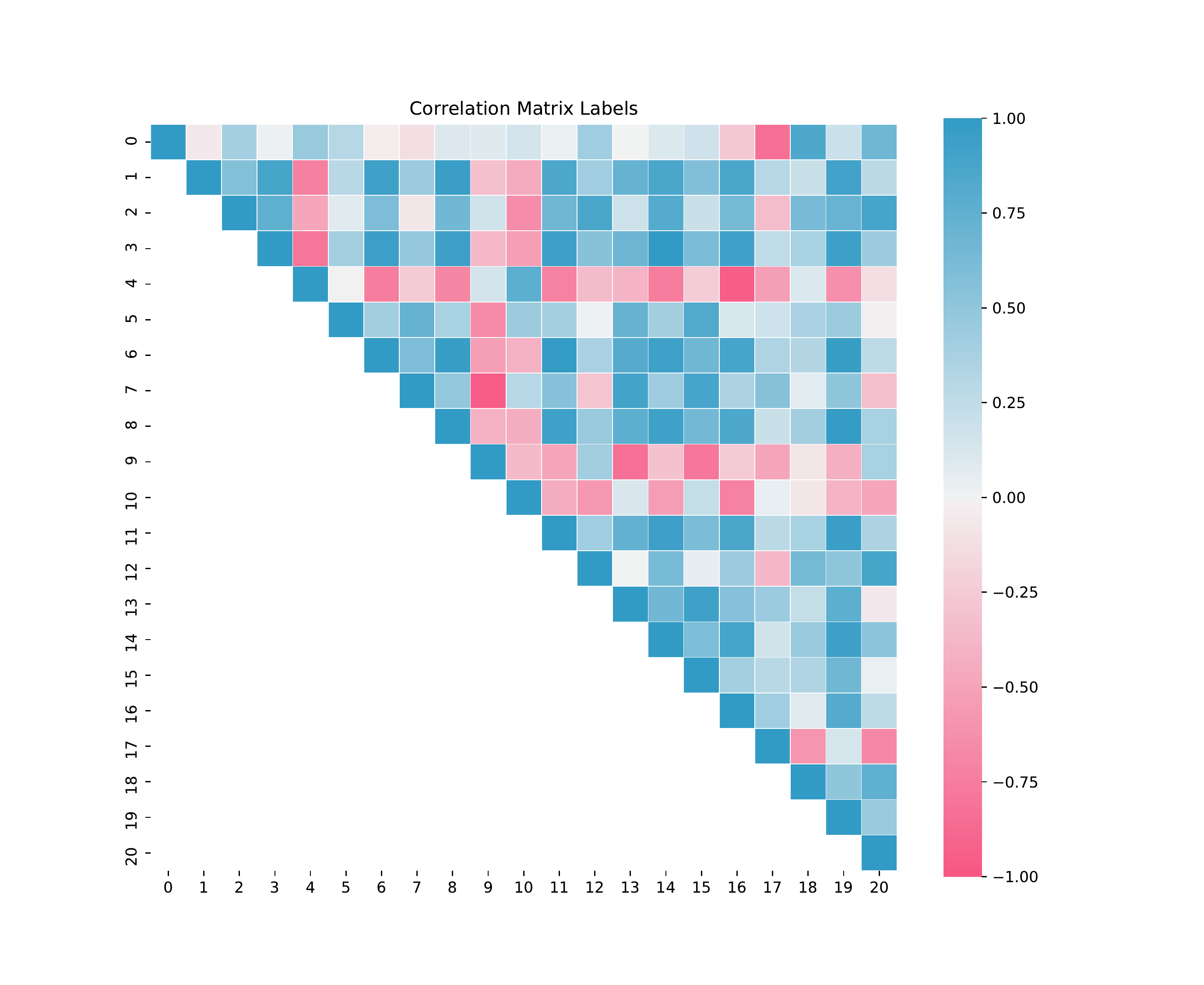}}\hfill
		\subfigure[MS-COCO, TBH]{
			\label{fig:correlation_labels_c}
			\includegraphics[width=0.32\textwidth]{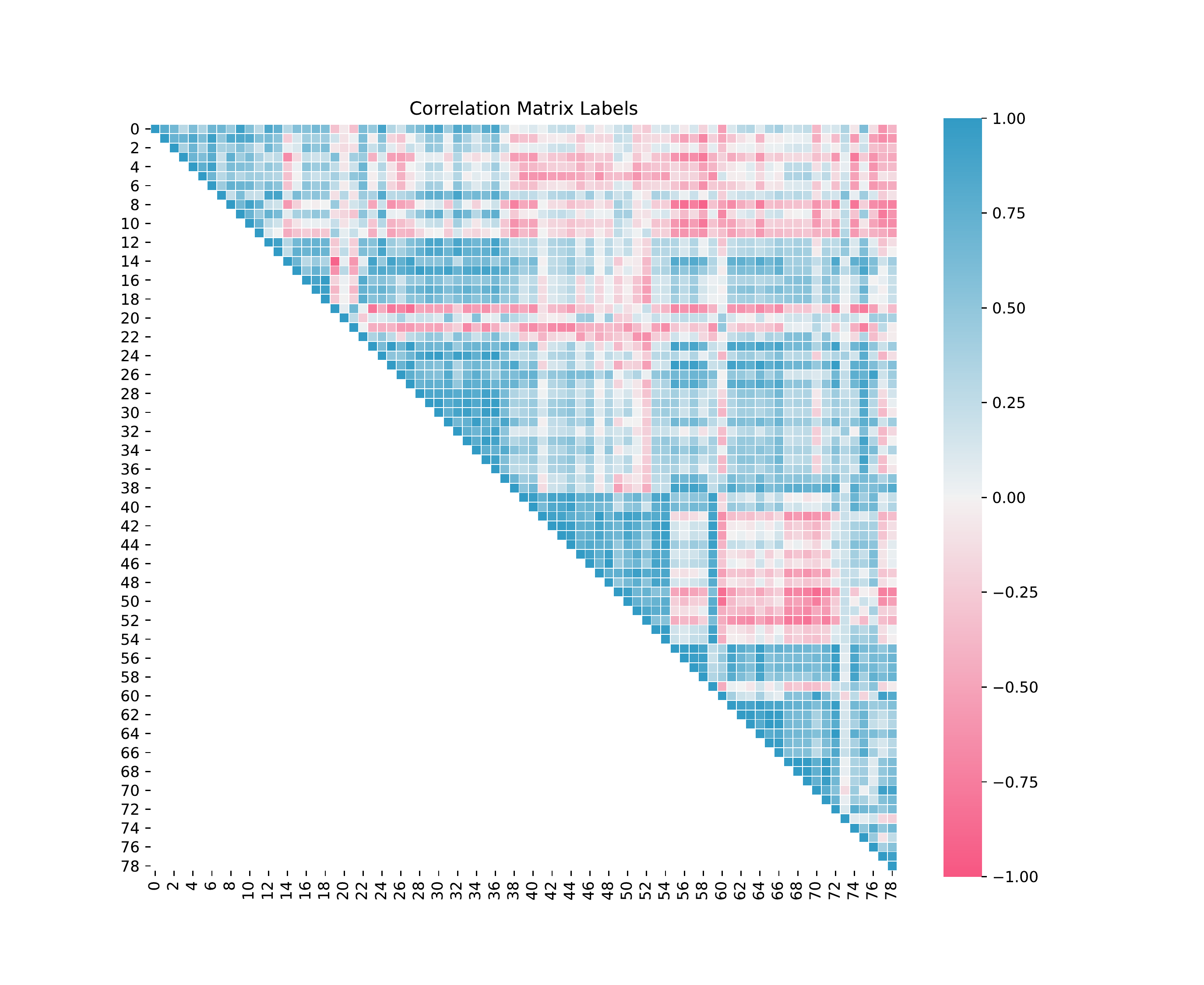}}\\
		\subfigure[CIFAR-10, STBH, $\eta=100,\gamma=20$]{
			\label{fig:correlation_labels_d}
			\includegraphics[width=0.32\textwidth]{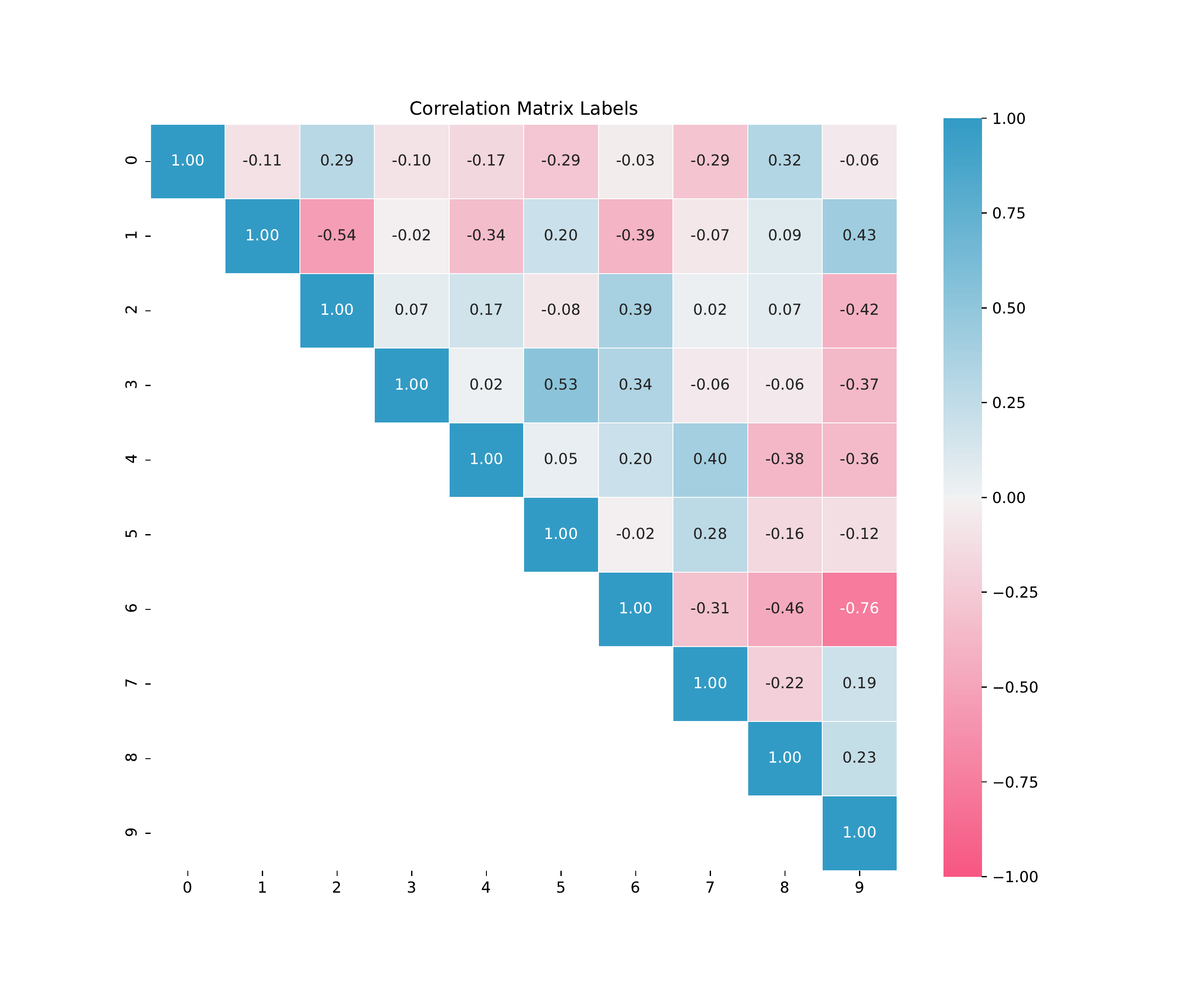}}\hfill
		\subfigure[NUS-WIDE, STBH, $\eta=20,\gamma=10$]{
			\label{fig:correlation_labels_e}
			\includegraphics[width=0.32\textwidth]{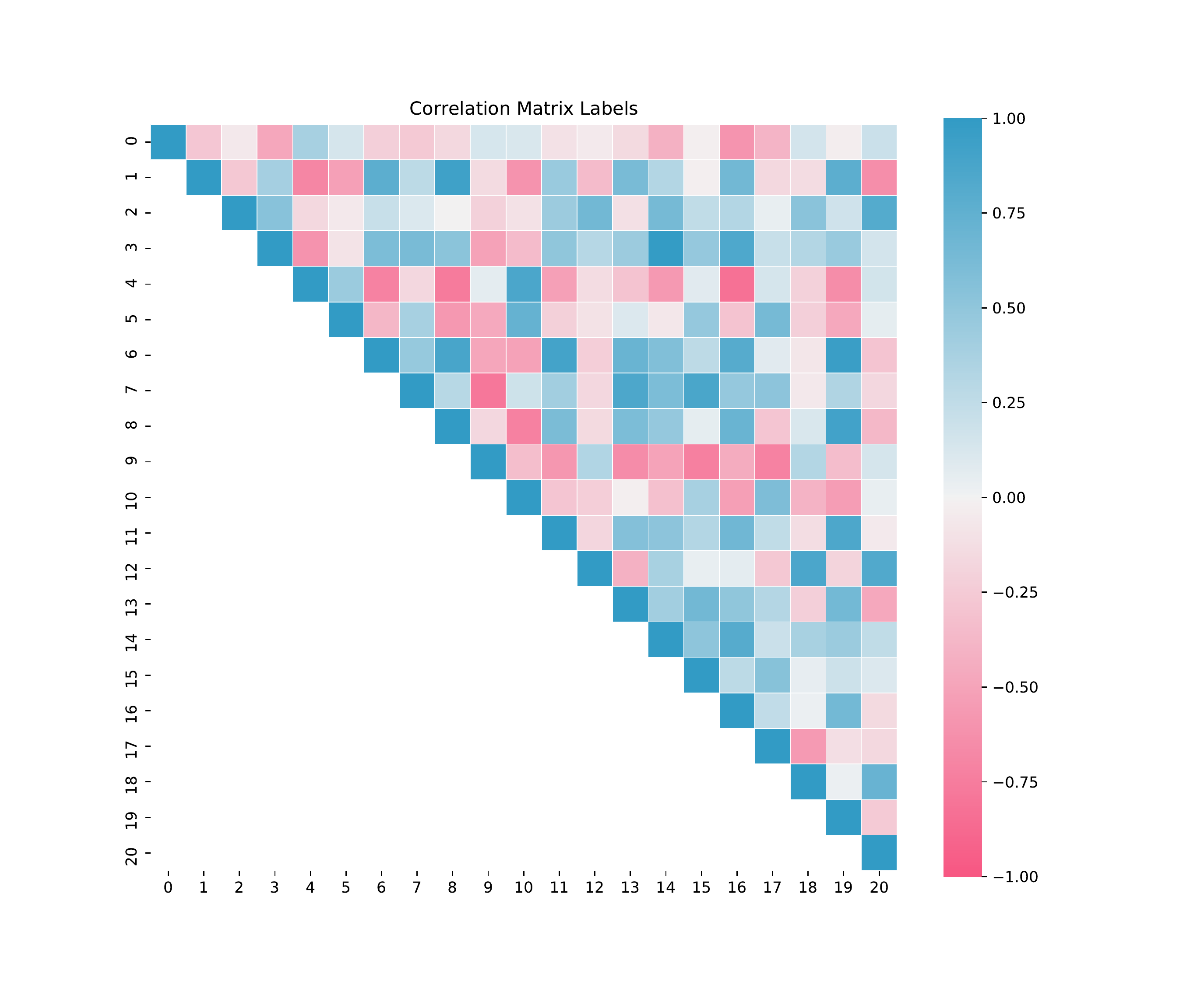}}\hfill
		\subfigure[MS-COCO, STBH, $\eta=30,\gamma=30$]{
			\label{fig:correlation_labels_f}
			\includegraphics[width=0.32\textwidth]{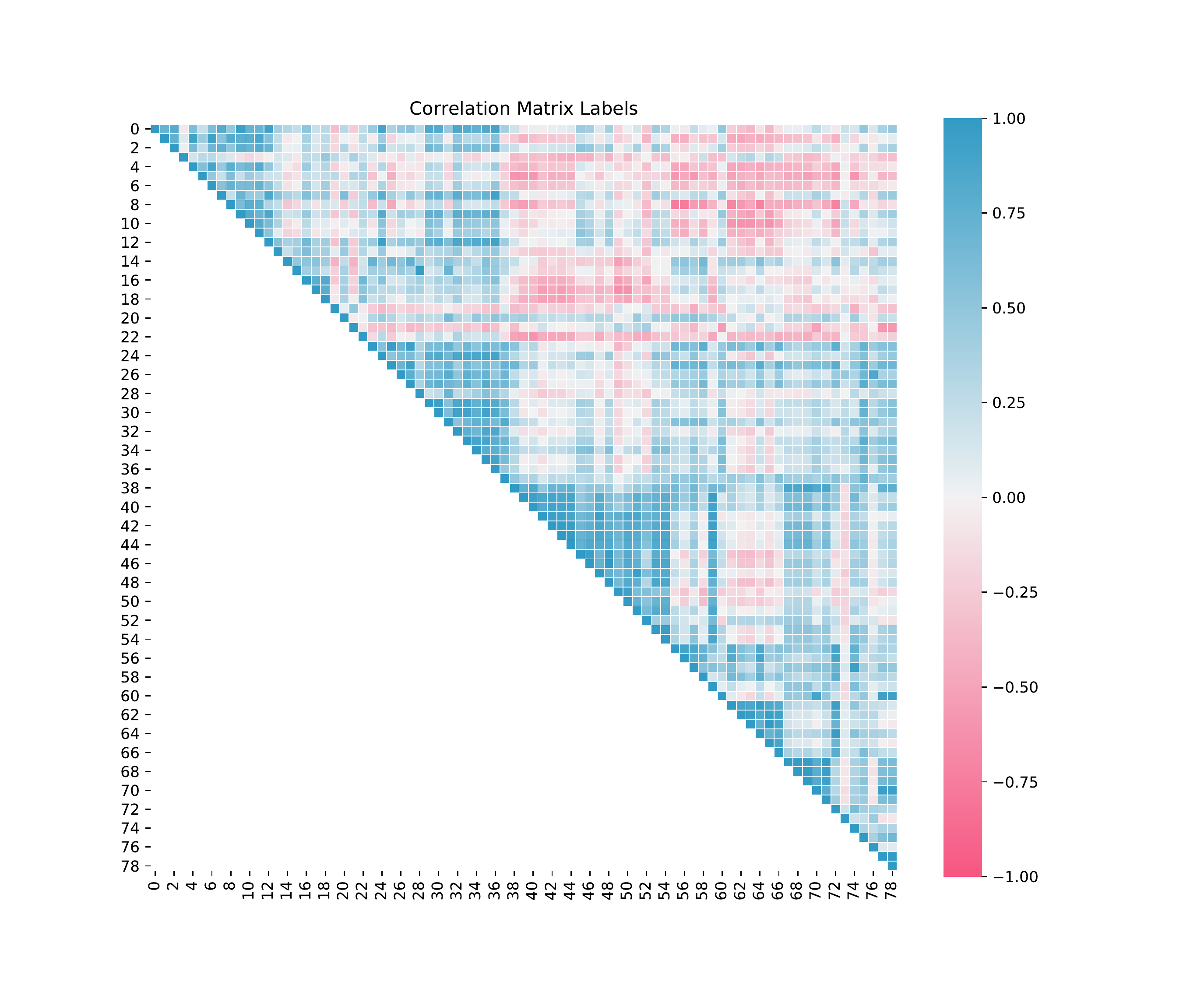}}
		\caption{ Correlation matrix of labels for the TBH model (upper) and STBH model (lower) with chosen $\eta$,$\gamma$ values. }
		\label{fig:correlation_labels}
	\end{figure*}

\begin{figure*}[!h]
		\centering 
		\subfigure[CIFAR-10, TBH]{
			\label{fig:correlation_hists_a}
			\includegraphics[width=0.32\textwidth]{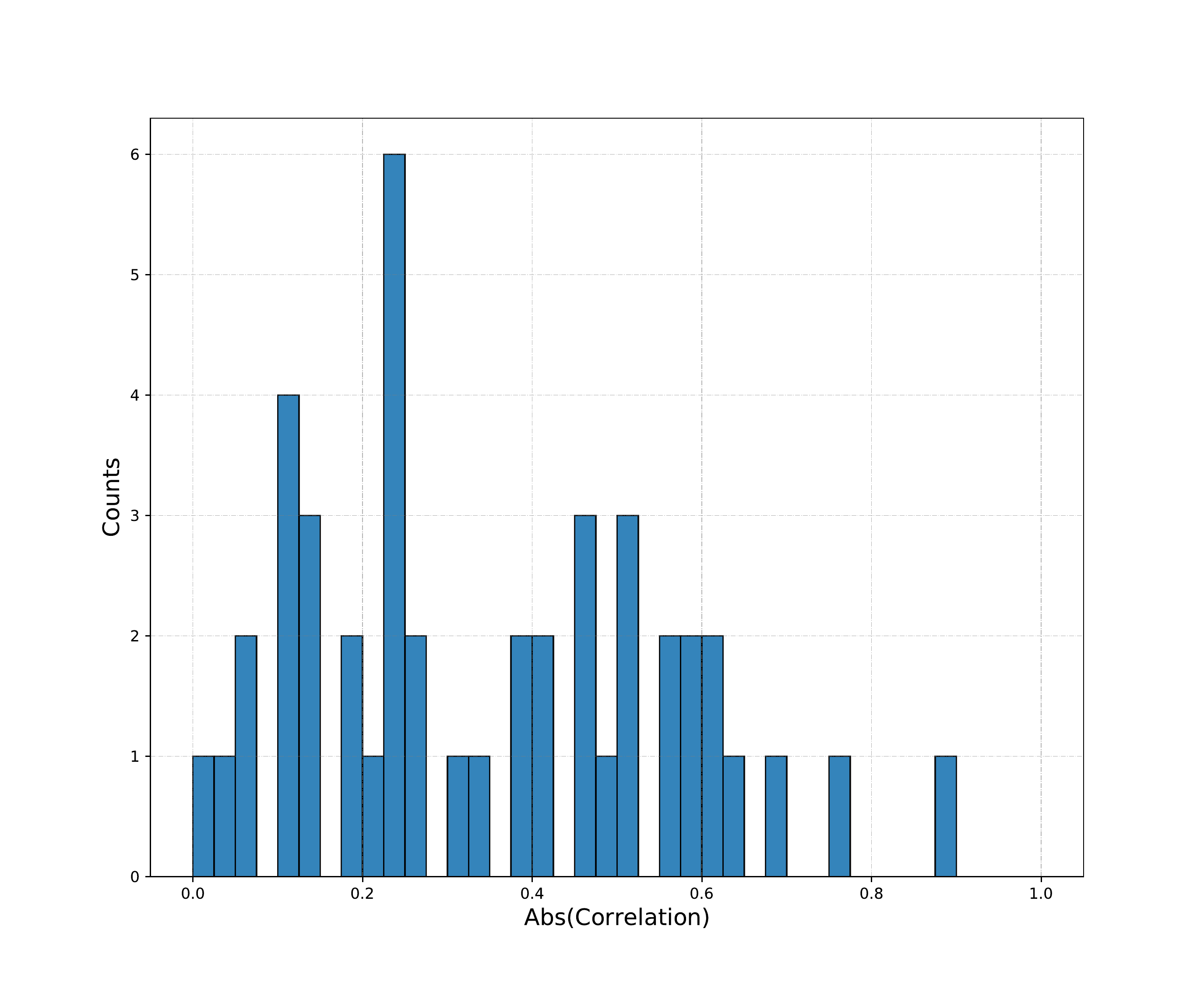}}\hfill
		\subfigure[NUS-WIDE, TBH]{
			\label{fig:correlation_hists_b}
			\includegraphics[width=0.32\textwidth]{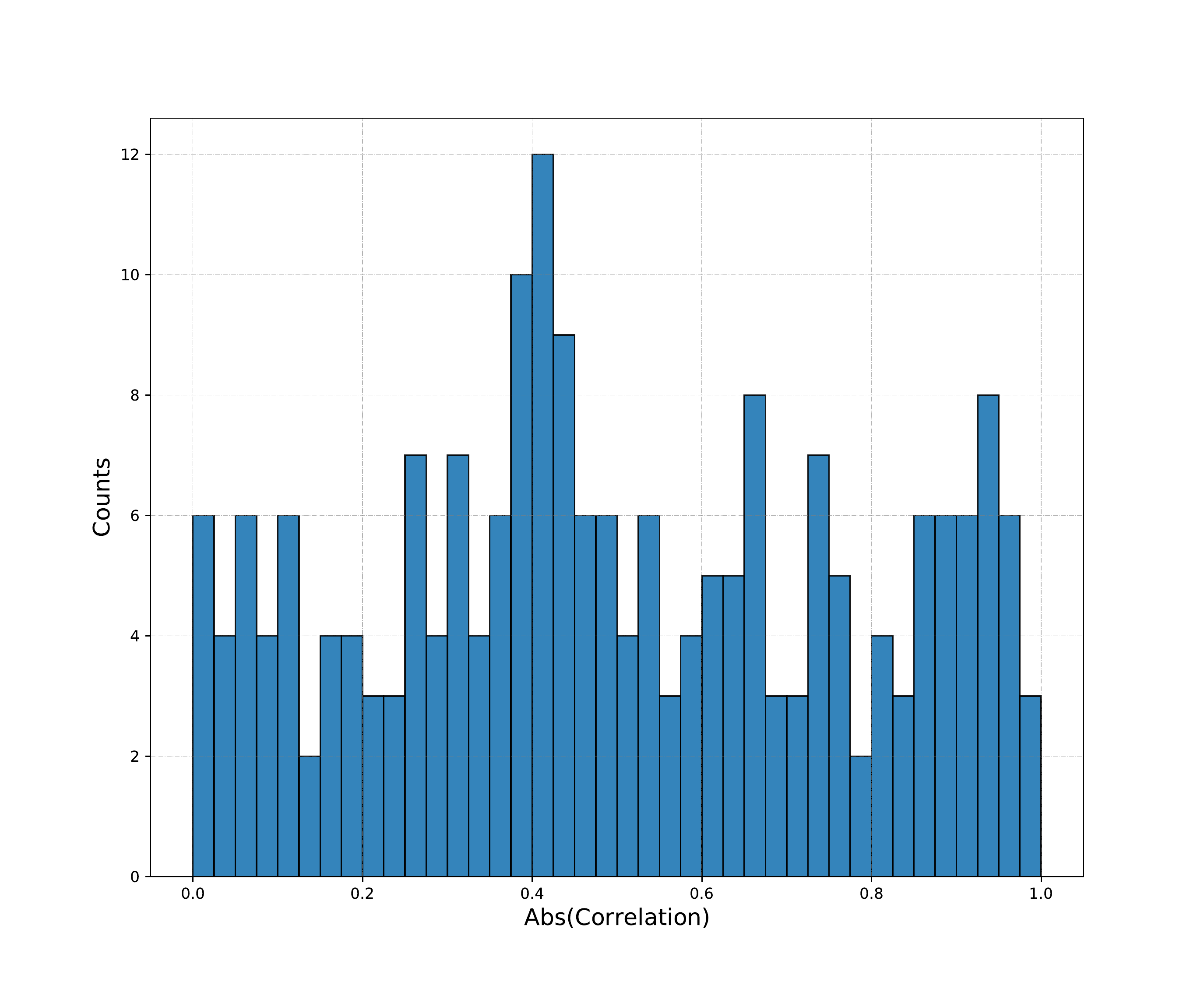}}\hfill
		\subfigure[MS-COCO, TBH]{
			\label{fig:correlation_hists_c}
			\includegraphics[width=0.32\textwidth]{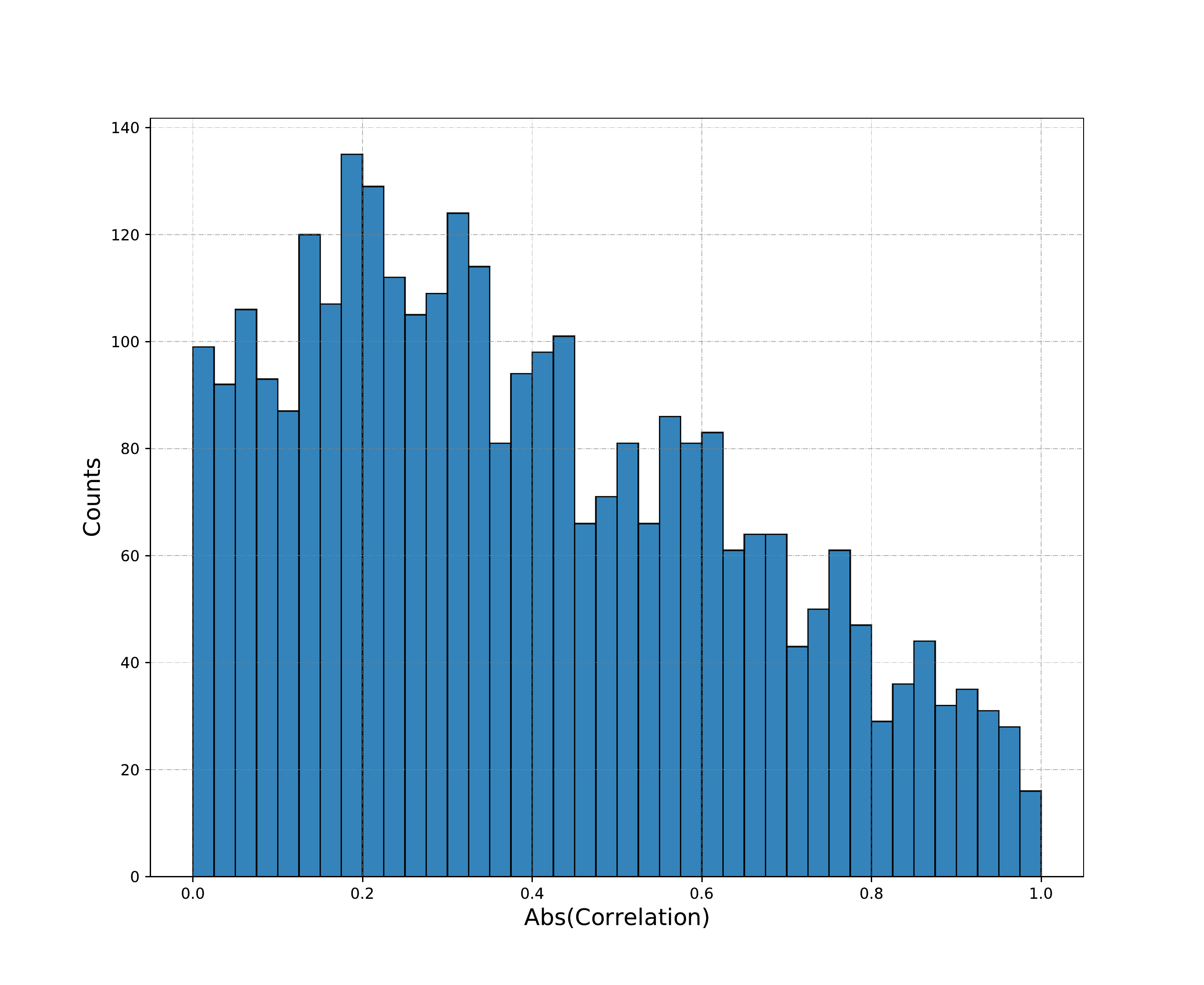}}\\
		\subfigure[CIFAR-10, STBH, $\eta=100,\gamma=20$]{
			\label{fig:correlation_hists_d}
			\includegraphics[width=0.32\textwidth]{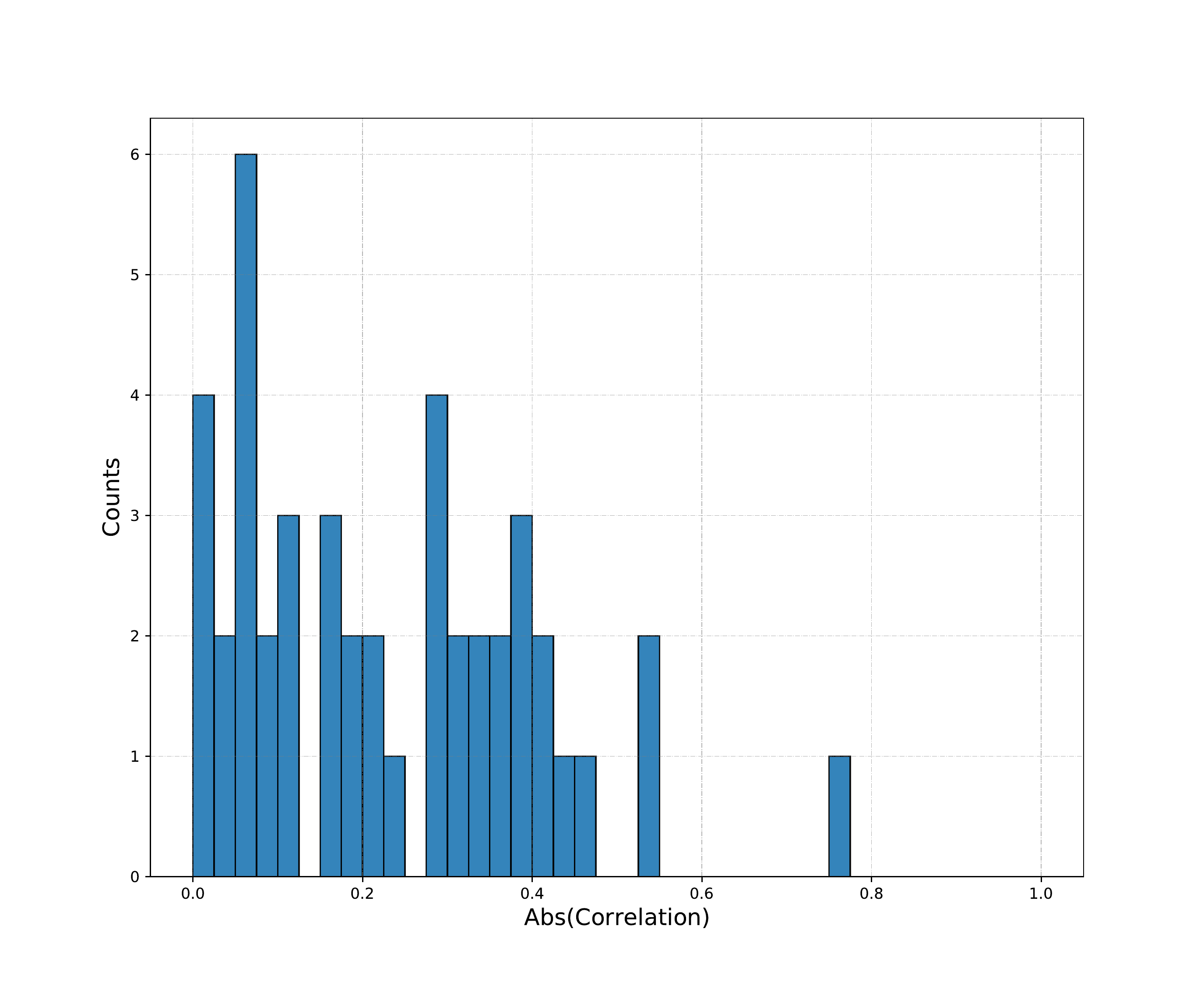}}\hfill
		\subfigure[NUS-WIDE, STBH, $\eta=20,\gamma=10$]{
			\label{fig:correlation_hists_e}
			\includegraphics[width=0.32\textwidth]{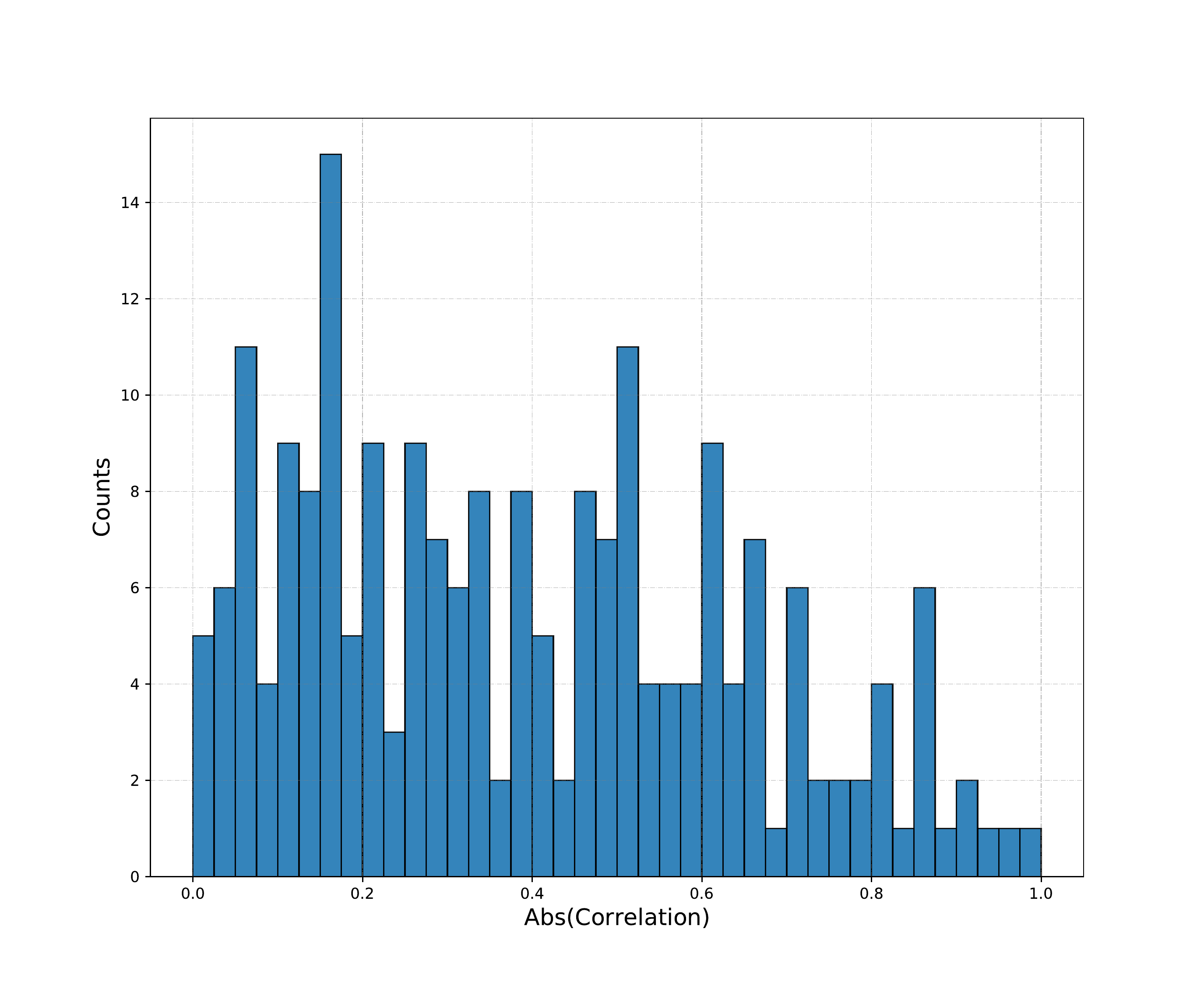}}\hfill
		\subfigure[MS-COCO, STBH, $\eta=30,\gamma=30$]{
			\label{fig:correlation_hists_f}
			\includegraphics[width=0.32\textwidth]{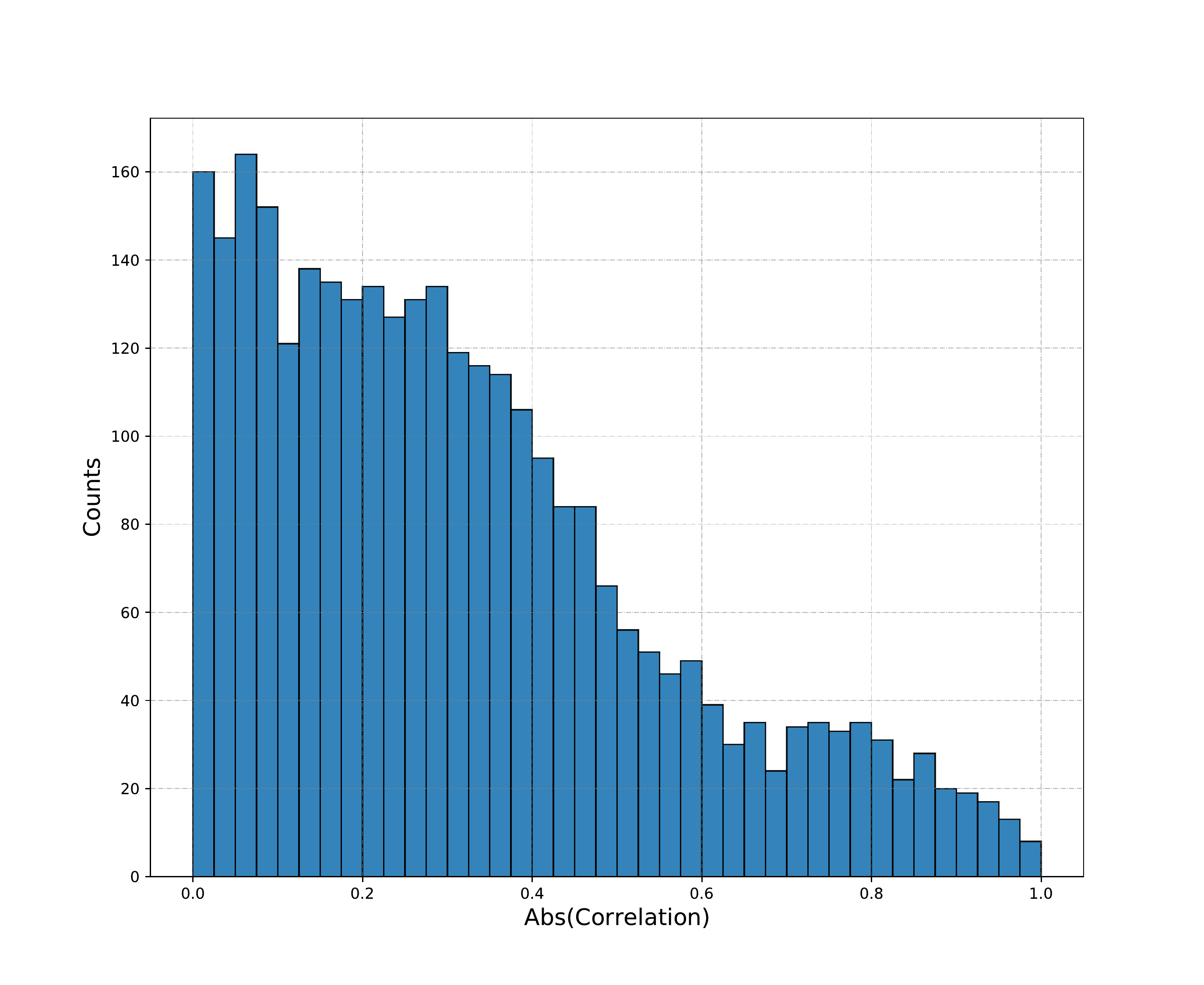}}
		\caption{ Histograms of absolute values of correlations for the TBH model (upper) and STBH model (lower) with chosen $\eta$,$\gamma$ values. }
		\label{fig:correlation_hists}
	\end{figure*}

	We have seen above that the Supervised TBH performs better than the original TBH model in terms of P@1000, P-R curve and mAP. As also mentioned, supervised learning could also help separate the codes' distributions for different labels since class labels can provide more accurate classification information. In other words, the correlation between different classes in STBH is expected to be lower than that of TBH. We show such plots of correlation matrices in Fig. \ref{fig:correlation_labels}, where we can see that the STBH (lower) generally has lower correlations. Note that for the STBH model, we choose the proper values of $\eta$ and $\gamma$ by grid searches for each of the three datasets. In Fig. \ref{fig:correlation_hists}, we take the absolute value of correlations, and plot the histograms of these numbers. Compared to the TBH model (upper), STBH (lower) mitigates strong correlations and shift the peak to the left. To quantify such differences, in Table \ref{tab:correlation_label}, we report the median, mean and standard deviation of these histograms between TBH and STBH models. From the table, we can again observe that the STBH helps untangle correlations and better discriminate between different labels.
	
	\begin{table*}[!h]
		\centering
		\fontsize{9}{12}\selectfont
		\caption{Median, mean and standard deviation of absolute values of the correlations for the TBH model and STBH model with the chosen $\eta$,$\gamma$ values.}
		\begin{tabular}{c|c|c|c|c|c|c}
			\toprule
			\multirow {2}{*}{\textbf{ }}&\multicolumn{2}{|c}{\textbf{CIFAR-10}} &\multicolumn{2}{|c}{\textbf{NUS-WIDE}}&\multicolumn{2}{|c}{\textbf{MS-COCO}} \cr
			\cmidrule(lr){2-3} \cmidrule(lr){4-5} \cmidrule(lr){6-7}
			\!&TBH\!&STBH,$\eta=100,\gamma=20$\!&TBH\!&STBH,$\eta=20,\gamma=10$\!&TBH\!&STBH, $\eta=30,\gamma=30$\!\cr
			\midrule
			
			\textbf{Median} & 0.349 & \textbf{0.203} & 0.461 & \textbf{0.347} & 0.353 & \textbf{0.275} \cr
			
			\textbf{Mean} & 0.301 & \textbf{0.232} & 0.501 & \textbf{0.388} & 0.395 & \textbf{0.319} \cr
			
			\textbf{Std} & 0.213 & \textbf{0.171} & 0.280 & \textbf{0.250} & 0.255 & \textbf{0.235} \cr
			\bottomrule
		\end{tabular}\label{tab:correlation_label}
	\end{table*}
	
	\subsection{Visualization}
	
	\begin{figure}[!h]
		\centering 
		\subfigure[t-SNE, TBH]{
			\includegraphics[width=0.48\linewidth]{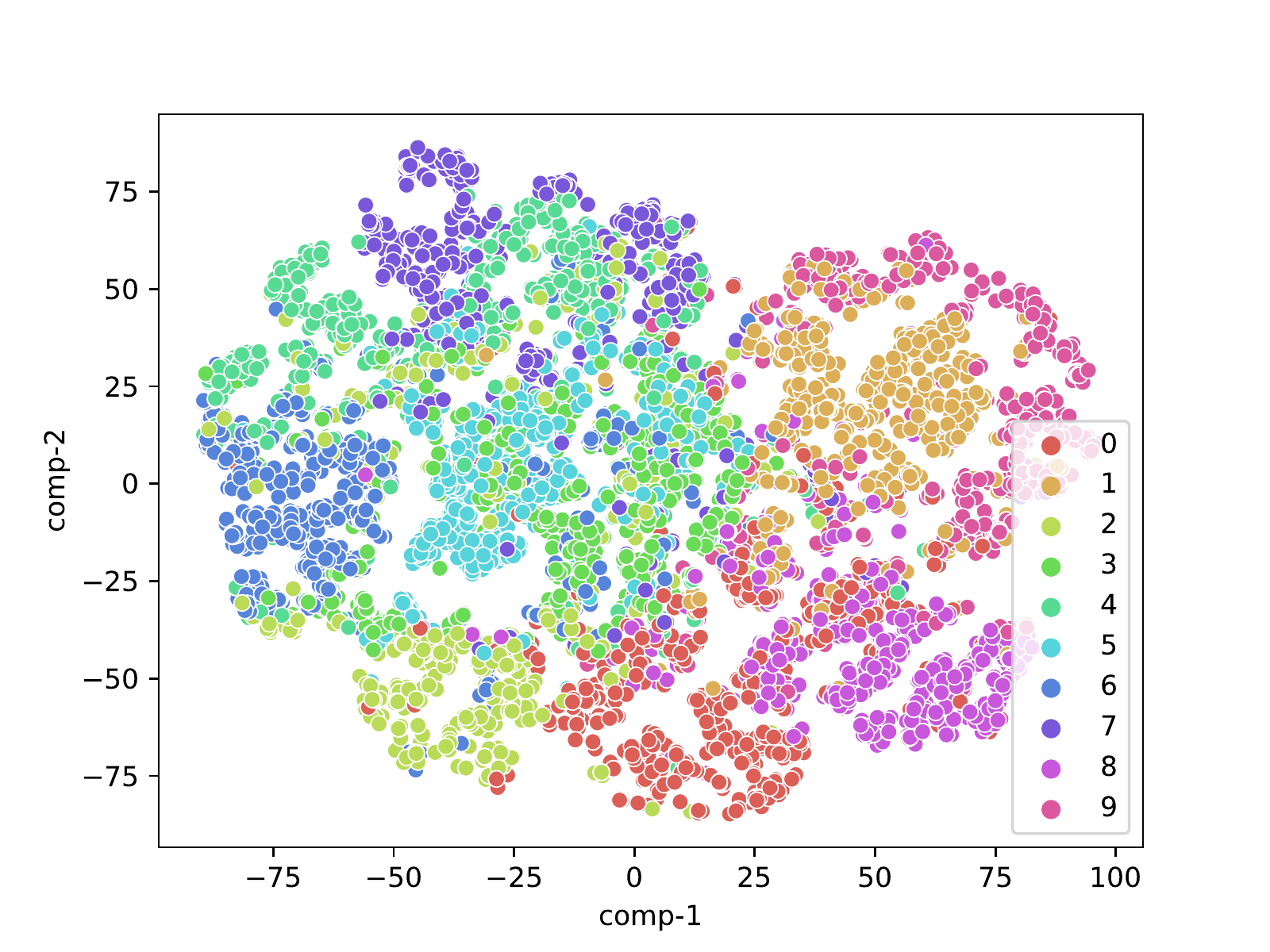}}\hfill
		\subfigure[t-SNE, STBH, $\eta=100,\gamma=20$]{
			\includegraphics[width=0.48\linewidth]{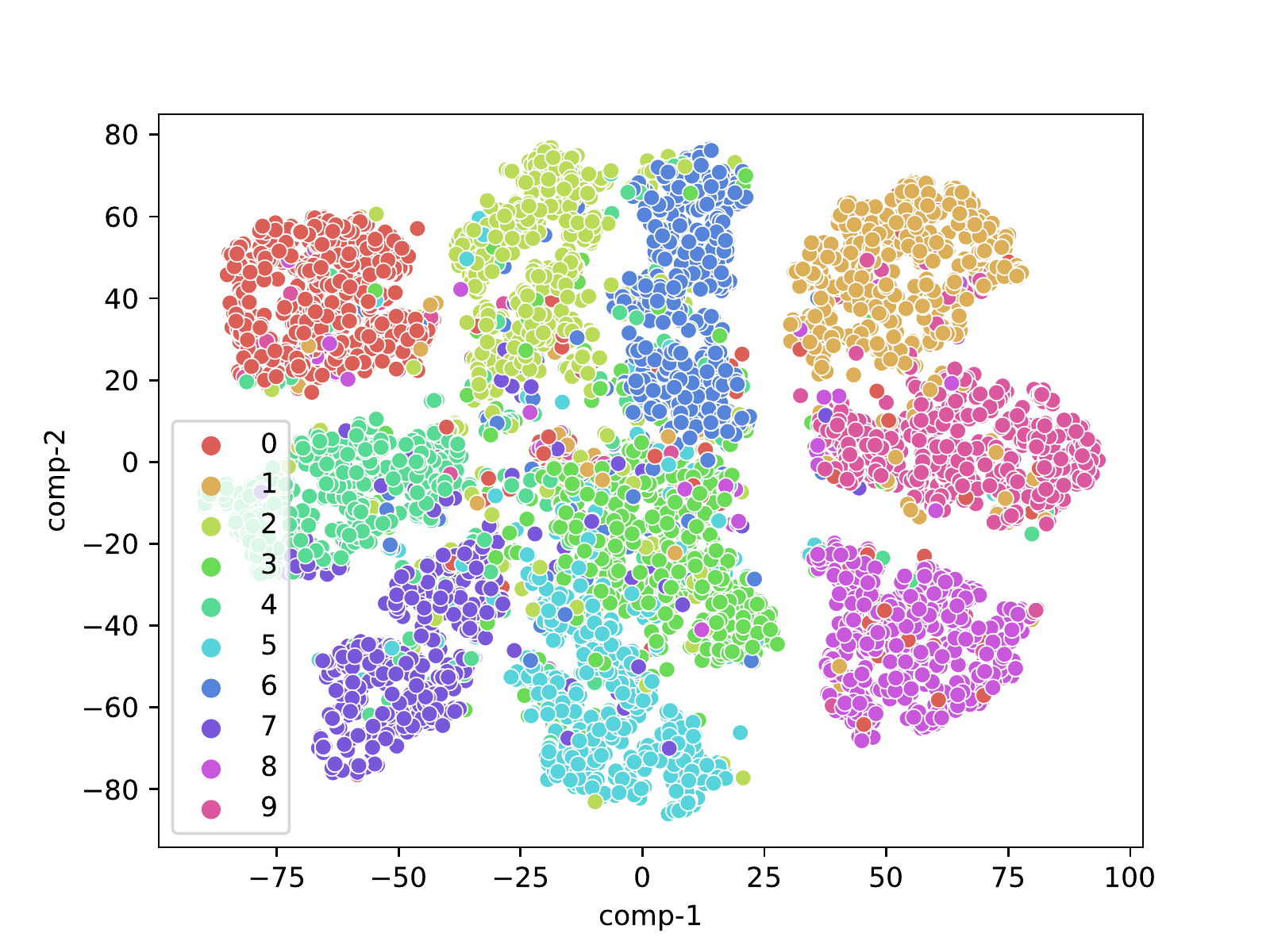}}\\
		\subfigure[UMAP,TBH]{
			\includegraphics[width=0.48\linewidth]{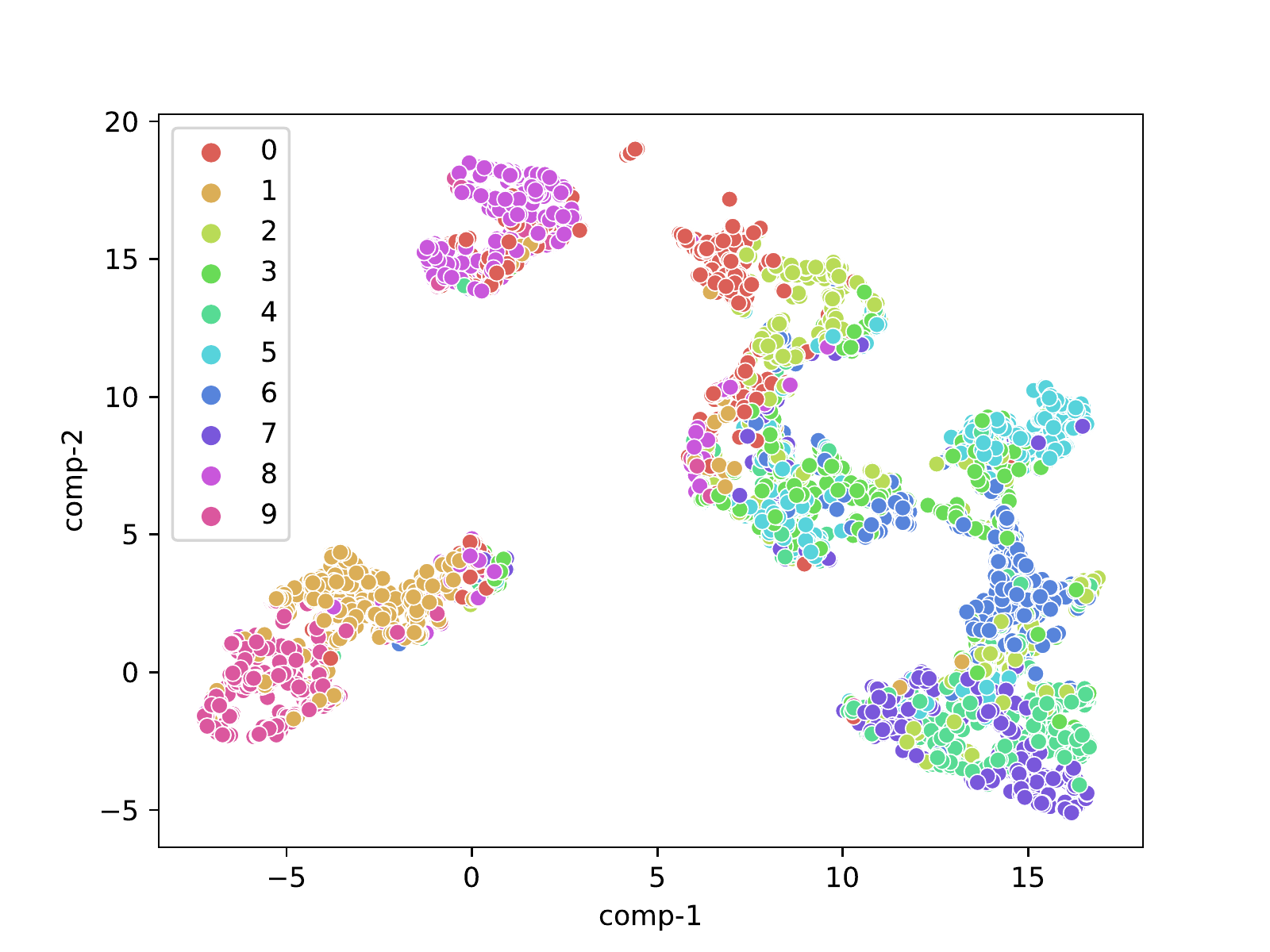}}\hfill
		\subfigure[UMAP, STBH, $\eta=100,\gamma=20$]{
			\includegraphics[width=0.48\linewidth]{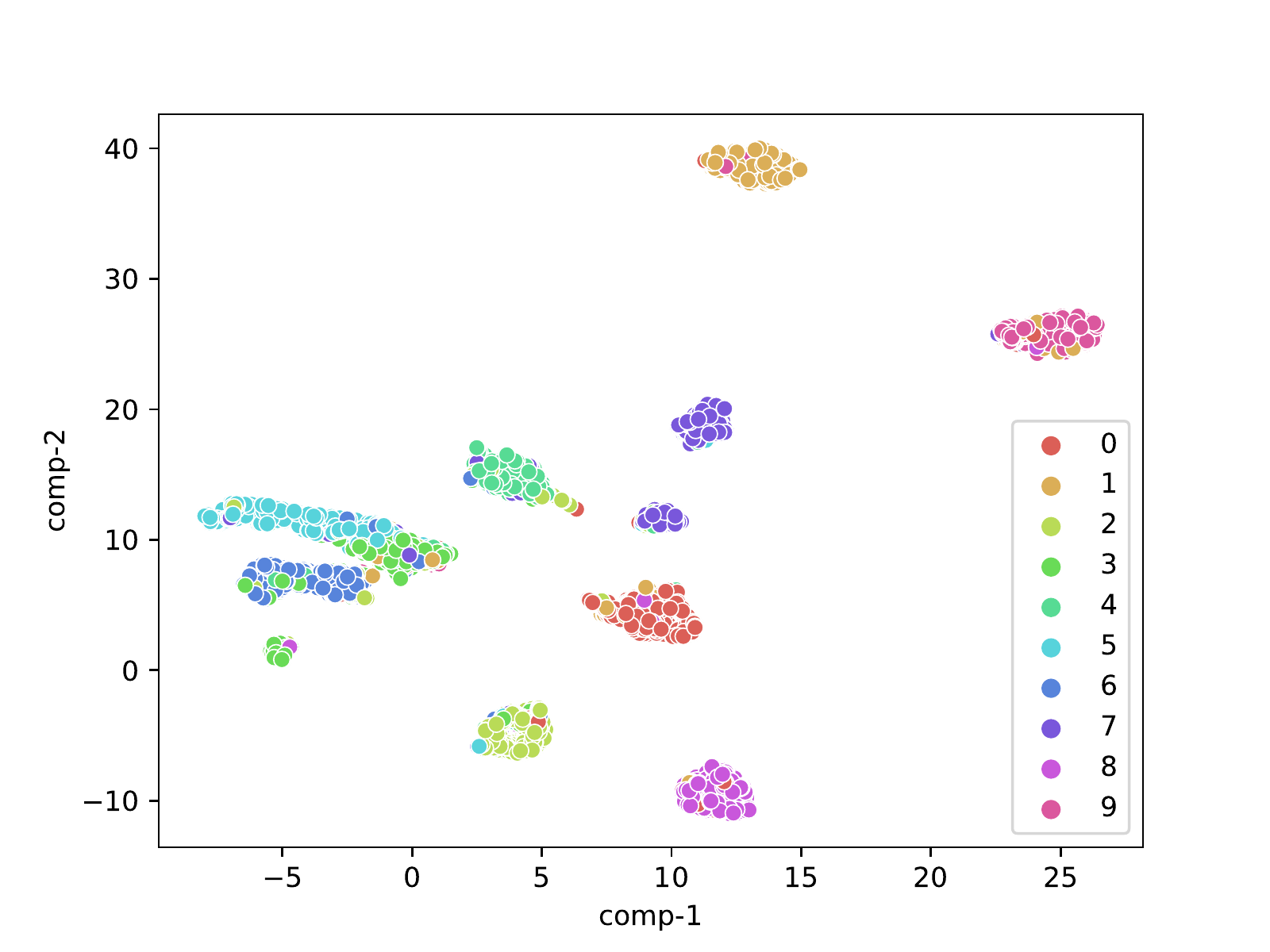}}
		
		\caption{ Visualization plots of CIFAR-10 from encoded binary codes with t-SNE (upper) and UMAP (lower) techniques for the TBH model (left) and STBH model (right). }
		\label{fig:visualization}
	\end{figure}
	
	To give an intuition on the discriminating power of the STBH model, Fig. \ref{fig:visualization} shows the t-SNE (upper) \cite{van2008visualizing} and UMAP (lower) \cite{mcinnes2018umap} visualization results on CIFAR-10 between TBH (left) and STBH (right) model. t-SNE is a popular technique to visualize high-dimensional data. UMAP is a novel algorithm for dimension reduction and visualization, but unlike t-SNE that pays more attention to the preservation of local distances, UMAP preserves more of the global structure of datasets \cite{mcinnes2018umap}. In Fig. \ref{fig:visualization}, we can clearly observe that STBH can better group data points of the same label together while pushing away samples from different labels with very clear boundaries. This illustrates the powerful discriminating ability of STBH due to the incorporated label information.
	
	\subsection{Comparison with Existing Supervised Methods}
	In the end, we also compare our STBH model with other existing state-of-art supervised deep hashing methods. For a fair comparison, we report results of binary codes with 32 bits for all methods, and numbers are adopted from their papers. Since we remove the "person" label from the MS-COCO dataset, MS-COCO is not included in the comparison. The mAP results are shown in Table \ref{tab:comp_other_SDH}, and the compared deep methods include: DTSH \cite{wang2016deep}, DPSH \cite{li2015feature}, DSDH \cite{li2017deep}, DRLIH \cite{peng2019deep}, PGDH \cite{yuan2018relaxation}, HBMP \cite{cakir2018hashing} and DRL-LER \cite{wang2020deep}. From Table \ref{tab:comp_other_SDH}, our results are comparable to the state-of-art performances, especially STBH outperforms all other methods on the NUS-WIDE dataset.

	\begin{table}[!h]
		\centering
		\fontsize{9}{12}\selectfont
		\caption{Performance comparisons of mAP with other Supervised Deep Hashing.}
		\begin{tabular}{c|c|c}
			\toprule
			\textbf{Methods} & \textbf{CIFAR-10} & \textbf{NUS-WIDE} \cr
			\midrule
			
			\text{DTSH} & 0.925 & 0.785 \cr
			\text{DPSH} & 0.795 & 0.736 \cr
			\text{DSDH} & 0.939 & 0.820 \cr
			\text{DRLIH} & 0.855 & 0.845 \cr
			\text{PGDH} & 0.741 & 0.780  \cr
			\text{HBMP} & 0.830 & 0.822 \cr
			\text{DRL-LER} & \textbf{0.952} & 0.837 \cr
			\hline
			\textbf{Ours} (\textbf{STBH}, $\eta=\gamma=50$) & 0.873 & \textbf{0.859} \cr
			\bottomrule
		\end{tabular}\label{tab:comp_other_SDH}
	\end{table}
	
	\section{Conclusion}
	
	In this paper, we propose the supervised twin-bottleneck hashing, where a classifier is applied on top of the binary bottleneck of the original TBH model \cite{shen2020auto}. The supervised learning enables the network to incorporate class labels, which provide complementary information independent of feature vectors, help separate data points with different labels, and mitigate strong correlations of classes. Experiments conducted on three representative datasets show a statistically significant improvement against the original TBH model at a small modification cost. Furthermore, STBH is compared to other supervised methods of deep hashing, which reveals our results are competitive to the state-of-art results.
	
	In addition, we briefly discuss the class imbalance problem that is hardly considered in previous hashing papers and also mention some existing possible solutions. More works can be performed to quantify the performance effect due to the imbalance, and deal with this problem in the future.
	
	\begin{appendices}
		
		\section{P-value computation in Hypothesis Testing}\label{sec:appendix_pvalue}
		\setcounter{equation}{0}
		\renewcommand{\theequation}{\thesection.\arabic{equation}}
		In the paper, we want to examine whether the STBH model performs better than the TBH model. More specifically, we would like to check if the mean of results from STBH ($\mu_{2}$) is greater than that from TBH ($\mu_{1}$). For this, we adopt the hypothesis testing with the following two hypotheses:
		\begin{equation}
			\begin{aligned}
				H_{0} &: \mu_{2} \leq \mu_{1} \\
				H_{1} &: \mu_{2} > \mu_{1}
			\end{aligned}
		\end{equation}
		where $H_{0}$ is the null hypothsis, $H_{1}$ is the alternative hypothesis, $\mu_{1}$ is the population mean of results from the TBH model, and $\mu_{2}$ is the population mean of the STBH model.\\
		\newline
		This is a question on the test of the difference between two population means. The distribution is assumed to be Gaussian. Considering the number of the samples is small (6 in our paper), we adopt the t-distribution $f_{\nu}(t)$ for the test statistic with $\nu$ the degree of freedom. And the test statistic $t^{*}$ is computed as follows:
		\begin{equation}
			t^{*} = \frac{ \left(\bar{X}_{1}-\bar{X}_{2} \right) }{ \left( \frac{s_{1}^{2}}{n_{1}} + \frac{s_{2}^{2}}{n_{2}} \right)^{1/2} }
		\end{equation}
		where $\bar{X}_{1}$ ($\bar{X}_{2}$) is the samples' mean of TBH (STBH), $s_{1}$ ($s_{2}$) is the samples' standard deviation of TBH (STBH), and $n_{1}$ ($n_{2}$) is the number of samples, which is 6 in our case. 
		In addition, the degree of freedom of the t-distribution is given by:
		\begin{equation}
			\mathtt{\nu} = \frac{ \left( \frac{s_{1}^{2}}{n_{1}} + \frac{s_{2}^{2}}{n_{2}} \right)^{2} }{ \frac{\left(s_{1}^{2}/n_{1} \right)^{2}}{n_{1}} + \frac{\left(s_{2}^{2}/n_{2}\right)^{2}}{n_{2}} }
		\end{equation}
		We can then compute p-value from the t-distribution with the computed test statistic as: 
		\begin{equation}
			p = \int_{t^{*}}^{\infty} f_{\nu}(t) dt
		\end{equation}
		The smaller the p-value is, the more evidence we have to reject the $H_{0}$ and accept the $H_{1}$ hypothesis.
		
	\end{appendices}
	
	{\small
		\bibliographystyle{ieee_fullname}
		\bibliography{sbth}
	}

\end{document}